\documentclass[sigconf, nonacm]{acmart}
\usepackage{booktabs}  
\usepackage{threeparttable}
\usepackage{multirow}
\usepackage{subfigure}
\usepackage{graphicx}
\usepackage{xcolor}
\usepackage{bm}
\usepackage[noend]{algpseudocode}
\usepackage{algorithmicx,algorithm}
\usepackage{times}
\usepackage{enumitem}
\usepackage{float}
\usepackage{mathtools}
\usepackage{array}
\usepackage{diagbox}
\usepackage[normalem]{ulem}
\usepackage{amsmath}

\newcommand\vldbdoi{10.14778/3503585.3503604}

\newcommand\vldbavailabilityurl{https://github.com/WXL520/AutoCTS}
\newcommand\vldbpagestyle{empty}

\begin{document}
\title{AutoCTS: Automated Correlated Time Series Forecasting -- Extended Version}

%


\author{Xinle Wu$^{1}$, Dalin Zhang$^{1}$, Chenjuan Guo$^{1}$, Chaoyang He$^{2}$, Bin Yang$^{1\ast}$, Christian S. Jensen$^{1}$}
\affiliation{%
  \institution{$^{1}$Aalborg University, Denmark \hspace{5pt} $^{2}$University of Southern California, USA}
  \institution{$^{1}$\{xinlewu, dalinz, cguo, byang, csj\}@cs.aau.dk \hspace{5pt} $^{2}$chaoyang.he@usc.edu}
}







\begin{abstract}

Correlated time series (CTS) forecasting plays an essential role in many cyber-physical systems, where multiple sensors emit time series that capture interconnected processes. Solutions based on deep learning that deliver state-of-the-art CTS forecasting performance employ a variety of spatio-temporal (ST) blocks that are able to model temporal dependencies and spatial correlations among time series. However, two challenges remain. First, ST-blocks are designed manually, which is time consuming and costly. Second, existing forecasting models simply stack the same ST-blocks multiple times, which limits the model potential. To address these challenges, we propose \emph{AutoCTS} that is able to automatically identify highly competitive ST-blocks as well as forecasting models with heterogeneous ST-blocks connected using diverse topologies, as opposed to the same ST-blocks connected using simple stacking. Specifically, we design both a micro and a macro search space to model possible architectures of ST-blocks and the connections among heterogeneous ST-blocks, and we provide a search strategy that is able to jointly explore the search spaces to identify optimal forecasting models. Extensive experiments on eight commonly used CTS forecasting benchmark datasets justify our design choices and demonstrate that \emph{AutoCTS} is capable of automatically discovering forecasting models that outperform state-of-the-art human-designed models. This is an extended version of ``AutoCTS: Automated Correlated Time Series Forecasting''~\cite{wupvldb}, to appear in PVLDB 2022.

\end{abstract}

\maketitle

\pagestyle{\vldbpagestyle}
\begingroup
\renewcommand\thefootnote{}\footnote{\noindent
$^*$: Corresponding author. \\
}\addtocounter{footnote}{-1}\endgroup

\section{Introduction}
We are witnessing continued developments in sensor technologies 
in 
cyber-physical systems (CPS), where sensors 
produce large amounts of correlated time series ~\cite{DBLP:conf/cikm/CirsteaMMG018,DBLP:journals/vldb/HuYGJ18}. For example, in transportation, traffic sensors embedded in roads emit multiple traffic time series that record traffic flows at their locations across time. 
Since the traffic on a road is often correlated with the traffic on nearby roads, the traffic time series are often correlated~\cite{DBLP:journals/pvldb/PedersenYJ20}. 
Forecasting on correlated time series plays an essential role in ensuring effective operation of CPSs, such as identifying trends, predicting future behavior~\cite{MileTS}, and detecting outliers~\cite{davidpvldb}. For instance, traffic time series forecasting can improve vehicle routing in transportation systems~\cite{DBLP:journals/vldb/GuoYHJC20,DBLP:conf/icde/LiuJYZ18,DBLP:journals/vldb/PedersenYJ20,tkdesean}.


By considering both \emph{temporal dependencies} in time series and \emph{spatial correlations} among different time series, recent deep learning models demonstrate impressive and state-of-the-art performance on correlated time series forecasting. 
Temporal dependencies capture how historical values influence future values. We use the term ``spatial correlations'' because the correlations among time series are often due to the proximity of the locations in which the sensors that generate the time series are deployed, but correlations may also be due to other factors. 
More specifically, correlated time series are modeled as a spatio-temporal (ST) graph, where nodes represent time series, and edges represent spatial correlations between pairs of time series~\cite{razvanicde2021, wu2020connecting}. 
%
%

\begin{figure}[htbp]
\center
\subfigure[Existing Model.
]{
\begin{minipage}[c]{\linewidth} 
\centering
\includegraphics[width=\linewidth]{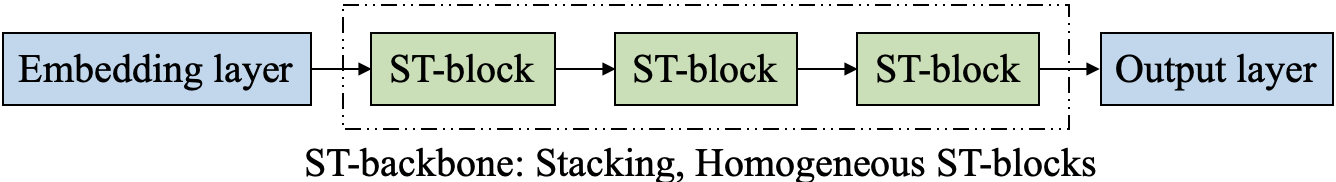}
\label{fig:existing st-model}
\end{minipage}
}
\subfigure[Example from the Proposed Model Space.
]{
\begin{minipage}[c]{\linewidth}
\centering
\includegraphics[width=\linewidth]{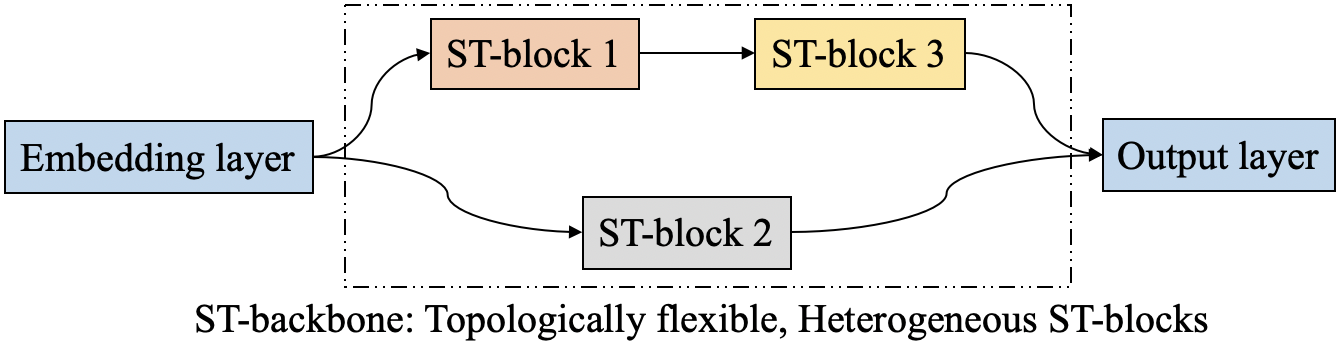}
\label{fig:proposed st-model}
\end{minipage}
}
\label{st-model}
\caption{
Existing vs. Proposed CTS Forecasting Models, with Different ST-blocks Colored Differently. 
}
\end{figure}

Based on the above ST-graph modeling, different models are proposed to enable forecasting. 
Figure~\ref{fig:existing st-model} summarizes existing forecasting models, which often 
include (1) an \emph{embedding layer} that transforms the input time series data, (2) a \emph{ST-backbone} that consists of a stack of multiple \emph{ST-blocks} that are able to extract appropriate spatio-temporal features from the embedded time series data, and (3) an \emph{output layer} that produces a final forecasting based on the features extracted by the ST-backbone.

Different studies propose unique {ST-blocks} that are responsible for the capture of both the temporal dependencies and spatial correlations~\cite{yu2018spatio,li2018dcrnn_traffic,guo2019attention,Wu2019graph,zheng2020gman,diao2019dynamic,fang2019gstnet,mengzhang2020spatial,xu2020spatial,MileTS,DBLP:conf/icde/Hu0GJX20}. 
%
%
For example, STGCN~\cite{yu2018spatio} employs a ``sandwich'' ST-block that includes two temporal convolutions, which model temporal dependencies, with one graph convolution in-between, which captures spatial correlations; and Graph Wavenet~\cite{Wu2019graph} uses a simpler ST-block that first employs gated temporal convolution to model temporal dependencies and then uses graph convolution to capture spatial correlations. 



Although various forecasting models have been proposed and the state-of-the-art accuracy continues to improve, two major limitations remain. 

\noindent \textbf{Manually Designed ST-blocks. }
Existing studies rely on human expertise to design effective ST-blocks, which is both inefficient, often taking weeks or months, and costly. 
%
In addition, as a large number of operators exist that are able to capture temporal dependencies (i.e., T-operators) and spatial correlations (i.e., S-operators), the search space for designing ST-blocks is very large. Thus, it is almost impossible for humans to be able to identify an optimal  combination of T/S-operators, thus potentially missing ST-blocks with high effectiveness. 
Further, rapid developments in machine learning are likely to lead to the invention of new, competitive operators. In order to benefit from new operators, it is necessary to reiterate the inefficient and costly manual design process whenever a new operator becomes available. Next, since time series from different domains have different characteristics, it is very difficult to design an ST-block that works well on substantially different datasets. 
To contend with these limitations, an automated design process that is able to identify optimal ST-blocks from a configurable search space of T/S-operators for specific datasets is called for. 


\noindent \textbf{ST-backbone with Stacking, Homogeneous ST-blocks.}
%
{Existing forecasting models often have an ST-backbone that stacks the same ST-blocks sequentially multiple times to achieve “deeper” models that are better at capturing complex dependencies and correlations, 
as shown in Figure~\ref{fig:existing st-model}. 
%
%
We hypothesize that ST-backbones with \emph{heterogeneous} ST-blocks connected by more \emph{flexible topologies}, 
as exemplified in Figure~\ref{fig:proposed st-model}, hold the potential to yield higher accuracy. Here, 
different ST-blocks, instead of same ST-blocks, can be 
connected using arbitrary topologies rather than by sequential stacking only. 
Intuitively, different ST-blocks may extract distinct features, which may enable more diverse and thus potentially better representations of time series. 
{Supporting multiple topologies} offers added flexibility, thus contributing further to enabling diverse models, which may enhance accuracy and stability. 
%
%
%
However, considering {topologically flexible}, heterogeneous ST-blocks increases the search space when designing forecasting models, thus rendering manual design  
more difficult and time-consuming. This naturally calls for an automated design approach. 
}

Although Neural Architecture Search (NAS), a technology that automatically learns neural architectures~\cite{elsken2019neural}, is able to outperform human-designed architectures on various tasks in computer vision (CV)~\cite{pham2018efficient,liu2018darts} and natural language processing (NLP)~\cite{wang2020textnas}, 
existing NAS methods fail to offer automated solutions capable of solving the aforementioned two limitations. 
First, no well-defined search space exists for correlated time series forecasting, as existing NAS methods often focus on CV and NLP. Directly using the search space designed for other domains fails to identify ST-blocks with high potential for capturing both temporal dependencies and spatial correlations. Directly using all existing S/T-operators in the literature yields an extremely large search space, thus making it very difficult and time-consuming to identify promising ST-blocks. 
Second, most existing NAS methods focus on identifying an optimal cell, e.g., an ST-block in our setting, while assuming a fixed topology, e.g., stacking the  same ST-blocks as shown in Figure~\ref{fig:existing st-model}, for connecting multiple instances of the same cell to  
%
derive the final model~\cite{elsken2019neural,zoph2018learning,pham2018efficient,liu2018darts}. {This fails to address the second limitation of manually designed forecasting models}. 
%
%

We propose \emph{AutoCTS} that is able to not only automatically design ST-blocks but also ST-backbones with complex topologies that connect heterogeneous ST-blocks, thus addressing the two limitations.  
%
%
To create \emph{AutoCTS}, we first design a micro search space targeting ST-blocks that models operators and how different operators are connected using a graph. To enable effective and efficient search, we 
judiciously select a compact set of T-operators that model temporal dependencies and S-operators that model spatial correlations based on a thorough analysis of existing, manually designed ST-blocks. 
This enables us to
automatically identify highly competitive ST-blocks in the proposed micro search space, thus addressing the first limitation.  
Next, we propose a macro search space, along with a 
joint search strategy that allows searches for an optimal topology among heterogeneous ST-blocks. 
This addresses the second limitation.


To the best of our knowledge, this is the first study that
systematically investigates automated correlated time series forecasting by exploring jointly the neural architectures of ST-blocks and ST-backbones. The study makes three contributions. 
First, we carefully design a micro search space for correlated time series forecasting, including both T-operators and S-operators, along with a search strategy that is able to identify optimal ST-blocks from the micro search space. 
Second, we propose a macro search space,  
along with 
a joint search strategy that 
searches both ST-blocks and forecasting models, while allowing flexible topologies among heterogeneous ST-blocks. 
Third, we conduct extensive experiments on correlated time series from different application domains to offer insight into and justify our design choices, demonstrating also that our proposal is able to outperform state-of-the-art methods. 


The remainder of the paper is organized as follows. Section~\ref{Sec 3} presents preliminaries. Section~\ref{Sec 4} elaborates the proposed \emph{AutoCTS} framework. Section~\ref{Sec 5} presents the empirical study. Section~\ref{Sec 2} covers related work, and  Section~\ref{Sec 6} concludes and suggests future work.

\begin{figure*}[htb]
  \centering
  \includegraphics[width=0.85\linewidth]{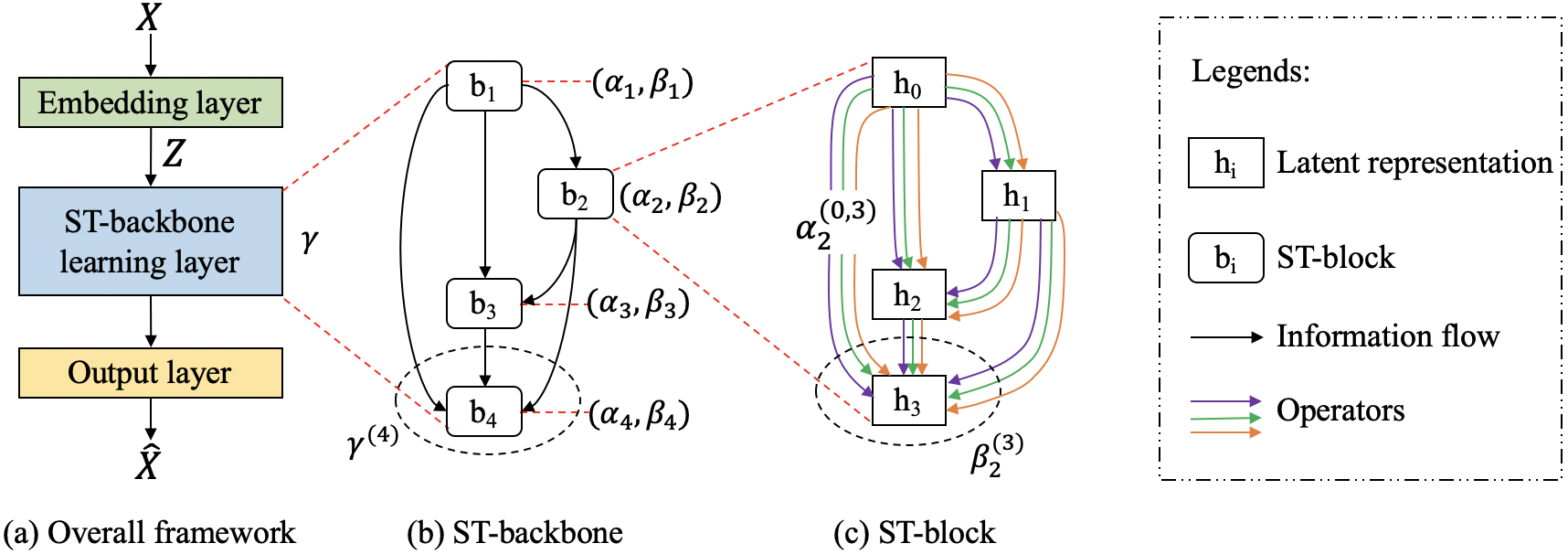}
  \caption{\emph{AutoCTS} Overview. 
  }
  \label{fig:framework}
\end{figure*}

\section{Preliminaries}
\label{Sec 3}
%
We introduce correlated time series forecasting, cover concepts that are necessary for the paper's proposal, and formalize the problem. 

\noindent\textbf{Correlated Time Series. }
Consider $N$ correlated multivariate time series
$\bm{\mathcal{X}} \in \mathbb{R}^{N\times T\times F}$, where each time series covers $T$ timestamps and each timestamp is associated with $F$ features. 
For example, assuming that 100 sensors are deployed in a road network and each sensor reports both travel speed and traffic flow every 5 minutes. Then, for one day, we have correlated time series $\bm{\mathcal{X}} \in \mathbb{R}^{100\times 288\times 2}$ with $N=100$ time series covering $T=288$ timestamps and $F=2$ features.  
%
We use $\bm{X}^{(i)} \in \mathbb{R}^{T\times F}$ to indicate the $i$-th time series, where $1\leq i\leq N$, and $\bm{X}_{t} \in \mathbb{R}^{N\times F}$ to indicate the features from all time series at timestamp $t$, where $1\leq t\leq T$. 
%

%


To model spatial correlations among different time series, we introduce 
a graph $G=(V,E,A)$, where each vertex in $V$ corresponds to a time series so that $|V|=N$, 
edges in $E$ represent spatial correlations between different time series, and adjacency matrix 
$A \in \mathbb{R}^{N\times N}$ contains edge weights that reflect the strengths of the spatial correlations between time series. The edge weights are either predefined, e.g., based on the distances between the locations of the sensors that generate the time series~\cite{li2018dcrnn_traffic,Wu2019graph,song2020spatial}, or learned in a data-driven manner~\cite{bai2020adaptive,wu2020connecting,razvanicde2021}. 

\noindent\textbf{Correlated Time Series Forecasting. }
We consider both single-step and multi-step correlated time series forecasting. 
Given the past $P$ steps, (1) for the single step forecasting, we predict the $Q$-th future step, where $Q\geq 1$; (2) for the multi-step forecasting, we predict a total of $Q$ future steps, with $Q>1$. 
Formally, we define the single-step correlated time series forecasting problem as follows:
\begin{align}
\bm{\hat{X}}_{{t+P+Q}}=\mathcal{F}_w(\bm{X}_{t+1}, \bm{X}_{t+2}, ..., \bm{X}_{t+P};G)
\label{3.1}
\end{align}
where $\mathcal{F}_w$ is a forecasting model and $w$ is its learnable parameters; and $\bm{\hat{X}}$ represents forecasted values. 
Likewise, the multi-step correlated time series forecasting problem is defined as follows:
\begin{align}
{\{\bm{\hat{X}}_{t+P+1}, \bm{\hat{X}}_{t+P+2}, ...,\bm{\hat{X}}_{t+P+Q}\}}=\mathcal{F}_{w}(\bm{X}_{t+1}, \bm{X}_{t+2}, ..., \bm{X}_{t+P};G)
\label{3.2}
\end{align}

\noindent\textbf{Problem Definition. }
The goal of the paper is to automatically identify an accurate forecasting model $\mathcal{F}_w$. This includes the identification of (1) architecture parameters $\theta$ that describe the model $\mathcal{F}$, e.g., which operators that are used in different ST-blocks and how the different ST-blocks are connected in the ST-backbone; and (2) model parameters $w$ that are used in the different operators, e.g., kernels in convolution operators and the projection matrices in attention operators.  {The objective function is show in Equation~\ref{3.2}.}
\begin{align}
\mathit{argmin}_{\theta, w}~ \mathit{ErrorMetric}(\mathcal{F}_w,\mathcal{D}),
\label{3.2}
\end{align}
where $\mathit{ErrorMetric}(\mathcal{F}_w,\mathcal{D})$ returns the forecasting error of the model $\mathcal{F}_w$ that is learned on a training dataset 
$\mathcal{D}$. %

\section{Automated CTS Forecasting}
\label{Sec 4}


Figure~\ref{fig:framework} offers an overview of the automated CTS forecasting framework \emph{AutoCTS}, which 
consists of three main components---an embedding layer, an ST-backbone learning layer, and an output layer. 

The embedding layer 
maps the original input feature from time series $X$ to a high-dimensional representation $Z$, which facilities extracting richer features from the input time series. 

The ST-backbone learning layer, which is the core component of \emph{AutoCTS}, is able to automatically design ST-backbones that encompass heterogeneous ST-blocks (as exemplified in Figure~\ref{fig:framework} (b)), where the design of the heterogeneous ST-blocks is also automated (as exemplified in Figure~\ref{fig:framework} (c)). 
When searching for ST-backbones, we use parameter $\gamma$ to parameterize the connections among different ST-blocks. 
For example, $\gamma^{(4)}$ controls how the three connections from ST-blocks $b_1$, $b_2$, and $b_3$ connect to ST-block $b_4$. 
When searching for ST-blocks, we search both (1) the operators between two representations, parameterized by $\alpha$, and (2) the different possible connections among different hidden representations, parameterized by $\beta$. For example, 
$\alpha_2^{(0,3)}$ represents the operators between hidden representations $h_0$ and $h_3$ in ST-block $b_2$, $\beta_2^{(3)}$ represents, in ST-block $b_2$, how the hidden representations $h_0$, $h_1$, and $h_2$ connect to the hidden representation $h_3$. 
We use unique sets of parameters $\{\alpha_i, \beta_i\}$ such that heterogeneous ST-blocks can be identified.  
%
The automatically designed ST-backbone takes as input the high-dimensional representation $Z$ from the embedding layer and extracts spatio-temporal features, which are fed to the output layer. 

Finally, the output layer 
makes the forecasting $\hat X$. We use a loss function, e.g., mean squared error, to measure the discrepancy between the forecast w.r.t. the ground truth to enable learning. 

In the following, we first identify an appropriate search granularity (in Section~\ref{ssec:searchgran}), then we introduce the design of a micro search space for ST-blocks (in Section~\ref{ssec:micro}) and a macro search space for ST-backbones (in Section~\ref{ssec:macro}). Finally, we present the search strategy that explores the micro and macro search spaces jointly to discover promising forecasting models (in Section~\ref{sec 4.3}).

\subsection{Search Granularity} 
\label{ssec:searchgran}

The search space can be constructed from operators of different granularities. A search space based on fine-granularity operators offers more flexibility and greater opportunities for identifying promising neural architectures that cannot be identified by human experts, but it often also yields a very large search space, the search of which takes prohibitively long time and requires excessive computational resources. In contrast, a search space based on coarse-granularity operators yields a smaller search space and thus speeds up the search process, but it may also introduce human biases that may prevent the identification of high-performance architectures. 

More specifically, in our problem setting, three different granularities exist. From coarse to fine, they are \textbf{ST-blocks}, \textbf{S/T operators}, and \textbf{basic computations}. 
We proceed to introduce the three granularities using a concrete example. Then, we discuss our design choices related to choosing the appropriate search granularity.

\begin{figure}[!htbp]
\center
\subfigure[ST-backbone]{
\begin{minipage}[c]{0.3\linewidth} 
\centering
\includegraphics[width=\linewidth]{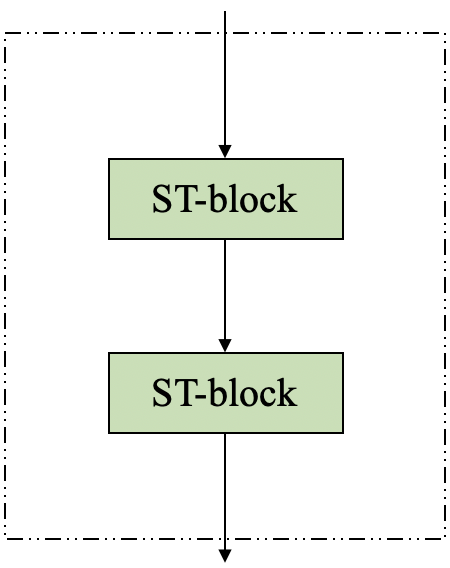}
\label{st-model}
\end{minipage}
}
\subfigure[ST-block]{
\begin{minipage}[c]{0.3\linewidth} 
\centering
\includegraphics[width=\linewidth]{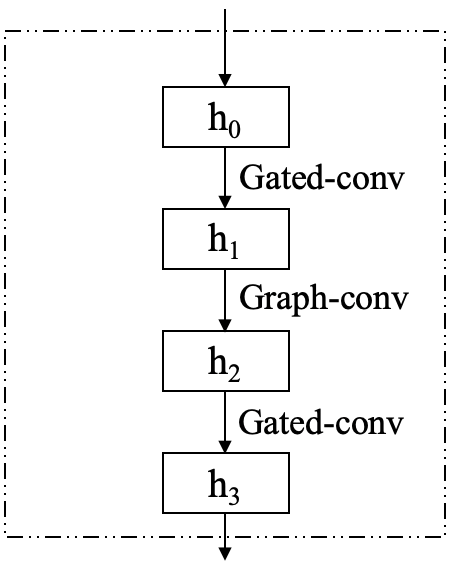}
\label{st-block}
\end{minipage}
}
\subfigure[Gated Convolution]{
\begin{minipage}[c]{0.3\linewidth} 
\centering
\includegraphics[width=\linewidth]{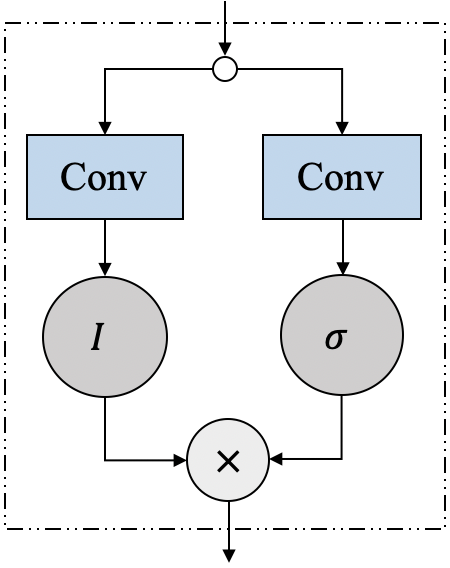}
\label{operator}
\end{minipage}
}
\caption{Human Designed Forecasting Model: STGCN.}
\label{granularity}
\end{figure}

{Figure~\ref{granularity} shows the neural architecture of Spatio-Temporal Graph Convolutional Networks (STGCN)~\cite{yu2018spatio}, a human designed forecasting model. The backbone of STGCN consists of two \textbf{ST-blocks} that are stacked (cf. Figure~\ref{st-model}). 
%
%
An ST-block consists of three \textbf{S/T operators}---two T-operators, i.e., gated convolution operators, 
with an S-operator in-between, namely a graph convolution operator, 
(cf. Figure~\ref{st-block}). %
An S/T operator often consists of multiple \textbf{basic computations}. 
For example, Figure~\ref{operator} shows the architecture of gated convolution, i.e., the T-operator.  
%
Here, $I$ and $\sigma$ refer to an identity and a sigmoid function, respectively; $Conv$ is the convolution operator, and $\times$ is the element-wise product. These are all basic computations. 
}

%

Based on the above, the coarsest search granularity is to use existing, manually-designed ST-blocks as atomic search units in the search space for finding novel ST-backbones. 
%
%
For example, instead of using a stacking structure with homogeneous ST-blocks as shown in Figure~\ref{st-model}, it is possible to search for an ST-backbone with a more flexible structure with many different human designed ST-blocks as shown in Figure~\ref{fig:proposed st-model}. 
%
%
However, since human designed ST-blocks may contain human biases already, only searching the different connections among them may limit the opportunities for 
finding novel and high-performance backbones. 


The next granularity is that of using, human designed S/T-operators as atomic search units to search for novel ST-blocks. 
Since S/T-operators are at a finer granularity than ST-blocks, this granularity offers greater opportunities for discovering more powerful forecasting models that go beyond existing human designed models. In addition, whenever a new S/T-operator is designed, 
the new S/T-operator can be easily included in the search space. We consider this as an appropriate granularity. 

The finest search granularity is to use basic operations as atomic search units to search for novel S/T operators. 
%
%
%
However, this leads to a much larger search space than when using S/T operators as the search space unit, incurs excessive computational costs,  and requires a very large dataset to enable effective training~\cite{mei2019atomnas}. 

To find highly competitive forecasting models without requiring high computational and memory costs, we choose to use S/T operators as the atomic search units in a so-called micro search space to discover novel ST-blocks. Next, in the macro search space, we use the automatically learned ST-blocks as atomic search units to identify novel ST-backbones with flexible structures. 



\subsection{Micro Search Space}
\label{ssec:micro}

The micro search space defines the possible architectures of the ST-blocks that can be discovered. 
We first introduce the design of the micro search space and then explain how to reduce the size of the micro search space to speed up search. 

\subsubsection{Micro-DAG}

We assume that an ST-block includes $M$ latent representations. 
The first latent representation is the output representation from the embedding layer or the output representation of another ST-block.
In addition, we consider a set $\mathcal{O}$ of operators, e.g., including multiple S/T operators, that are able to transform one latent representation to a new latent representation.  
%

\begin{figure}[!htbp]
\center
\subfigure[Micro-DAG]{
\begin{minipage}[c]{0.35\linewidth} 
\centering
\includegraphics[width=\linewidth]{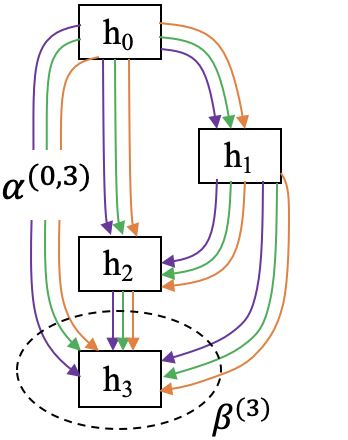}
\label{micro}
\end{minipage}
}
\subfigure[Derived ST-block]{
\begin{minipage}[c]{0.35\linewidth} 
\centering
\includegraphics[width=\linewidth]{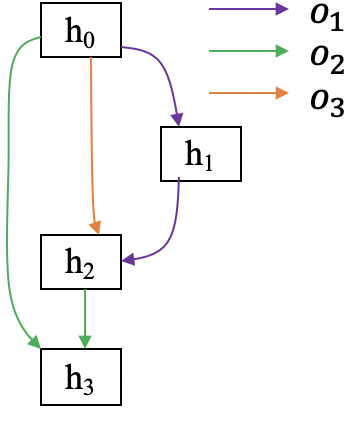}
\label{micro-derive}
\end{minipage}
}
\caption{Micro Search Space.}
\label{fig:micro}
\end{figure}

We represent the micro search space as a directed acyclic graph, denoted as micro-DAG (see Figure~\ref{fig:micro}).
%
%
%
The micro-DAG has $M$ nodes $h_i$,  $0\leq i\leq M-1$, that each denotes a latent representation. 
Node $h_0$ denotes the representation returned by the embedding layer. 
For each node pair $(h_i, h_j)$, we have $|\mathcal{O}|$ edges, where each edge corresponds to an operator from operator set $\mathcal{O}$. 
%
%
Figure~\ref{micro} shows an example, where $\mathcal{O}=\{o_1, o_2, o_3\}$ includes three operators such that each node pair is associated with three edges. 
%
In addition, we only include edges from node $h_i$ to $h_j$ if $i<j$. This makes the graph a DAG, which simulates the forward flow when training a neural network. 

%
Figure~\ref{micro-derive} shows a derived ST-block, which is a subgraph of the micro-DAG. Specifically, the derived ST-block only retains one edge, i.e., one operator, between each node pair $(h_i, h_j)$. In addition, for each node, it preserves at most two incoming edges. This enables relatively complex internal topologies for ST-blocks and avoids introducing too many parameters. 

The micro-DAG represents all possible architectures of an ST-block with $M$ latent representations. 
This design yields to  $|\mathcal{O}|^{\frac{M(M-1)}{2}}$ possible ST-blocks. 
This is because a micro-DAG with $M$ nodes has $\frac{M(M-1)}{2}$ node pairs $(h_i, h_j)$, where $i<j$, and because each node pair can be connected by an operator from $\mathcal{O}$. 
%
In Section~\ref{sssec:reduce}, we discuss how to select a compact operator set $\mathcal{O}$, thus reducing the size of the micro search space without comprising effectiveness.


\subsubsection{Parameterizing ST-blocks}
\label{ssec:parastblocks}
In order to derive an optimal ST-block, we introduce two sets of architecture parameters ${\alpha}$ and ${\beta}$, where ${\alpha}$ parameterizes node pair and $\beta$ parameterizes nodes. 

First, we parameterize each node pair $(h_i, h_j)$ with 
vector $\alpha^{(i,j)}\in \mathbb{R}^{|\mathcal{O}|}$ to indicate the weights over all operators in $\mathcal{O}$. 
Then {transformation $f^{(i,j)}$} from node $h_i$ to node $h_j$ is formulated as a weighted sum of all operators. 
\begin{align}
{f^{(i,j)}}=\sum\limits_{o\in \mathcal{O}}\frac{exp(\alpha_o^{(i,j)})}{\sum_{o'\in \mathcal{O}}exp(\alpha_{o'}^{(i,j)})}o(h_i), 
\label{4.2.1}
\end{align}
where $\alpha_o^{(i,j)}$ represents the weight of 
operator $o\in\mathcal{O}$, which is to be learned, and $o(h_i)$ is the representation after applying operator $o$ to representation $h_i$. 

Next, we parameterize each node $h_j$ based on its incoming edges. 
%
We use another architecture parameter $\beta^{(j)} \in \mathbb{R}^{j}$ to assign weights to the incoming edge groups at node $h_j$, where each incoming edge group represents a hidden representation from a node $h_i$ that appears earlier than $h_j$, i.e., $0\leq i< j-1$. 
For example, in Figure~\ref{fig:micro}, node $h_3$ has three incoming edge groups from $h_0$, $h_1$, and $h_2$, respectively. 
%
We then apply a softmax function to normalize the  ${\beta}$ parameter. If $SoftMax(\beta^{(3)})=(0.3, 0.3, 0.4)$, it means that the weights of the representations from $h_0$, $h_1$, and $h_2$ are 0.3, 0.3, and 0.4, respectively.
Therefore, for each node $h_j$, we can compute its representation as the weighted sum of all transformations of its predecessor nodes. 
\begin{align}
h_j&=\sum\limits_{i<j}\frac{exp(\beta^{(j)}[i])}{\sum_{i<j}exp(\beta^{(j)}[i])} 
{f^{(i,j)}} \\
&=\sum\limits_{i<j}\frac{exp(\beta^{(j)}[i])}{\sum_{i<j}exp(\beta^{(j)}[i])}\sum\limits_{o\in \mathcal{O}}\frac{exp(\alpha_o^{i,j})}{\sum_{o'\in \mathcal{O}}exp(\alpha_{o'}^{i,j})}o(h_i),
\label{4.2}
\end{align}
where $\beta^{(j)}[i]$ is the architecture parameter value for weighting the transformation from $h_i$ to $h_j$, which is to be learned. 

In this way, given the first node $h_0$, 
we are able to compute the representations of the remaining nodes in the micro-DAG. 
We use the representation of the last node $h_{M-1}$ as the output of the micro-DAG, and we thus feed $h_{M-1}$ to the output layer. This gives a forecasting model that we can train using classic back propagation. The training enables us to identify the most appropriate $\alpha$ and $\beta$. 


After training, we 
derive the final ST-block. 
For each node pair $(h_i, h_j)$, we compute a weight $w_o^{(i,j)}$ using Eq.~\ref{eq:wo} for each operator $o$, and retain the operator with the largest $w_o^{(i,j)}$, i.e., $argmax_{o\in \mathcal{O}} w_o^{(i,j)}$. 

\begin{equation}
\label{eq:wo}
w_o^{(i,j)}=\frac{exp(\beta^{(j)}[i])}{\sum_{i<j}exp(\beta^{(j)}[i])} \frac{exp(\alpha_o^{i,j})}{\sum_{o'\in \mathcal{O}}exp(\alpha_{o'}^{i,j})}
\end{equation}

Next, for each node $h_j$, we preserve two operators. One is the operator from node $h_{j-1}$, i.e., its immediate predecessor node. The other one is the operator with the largest $w_o^{(i,j)}$ among the remaining operators, i.e., $argmax_{0 \leq i \leq j-2} w_o^{(i,j)}$. For example, in Figure~\ref{micro-derive}, $h_{i-1}$ always connects to $h_{i}$, where $1 \leq i  \leq 3$. For $h_3$, assuming $argmax_{0 \leq i \leq 1} w_o^{(i,j)}=0$, then $h_0$ is connected to $h_3$ . 

\begin{table*}[h]
\small
    \centering
    \caption{Categorization of S/T Operators for Correlated Time Series Forecasting. 
    }
    \begin{threeparttable} 
    \begin{tabular}{l|l|l|l|l}
        \toprule  
        &\emph{Family}&\emph{Operator}&\emph{Literature}&\emph{Equation} \cr
        \hline
        \multirow{6}{*}{\rotatebox{90}{\emph{T-Operators}\phantom{Invisible}}}&\multirow{2}{*}{\textbf{CNN}}&1D Convolution&\cite{guo2019attention}&
        $H^{(i)}=Z^{(i)}\ast W
        $
        \begin{minipage}{1cm}\begin{equation}
            \hspace{5pt}
            \label{eq:1DC}
        \end{equation}\end{minipage}
            
        \cr

    \cline{3-5}
    &&\textbf{Gated Dilated Causal Convolution}&\multirow{2}{*}{\cite{yu2018spatio,diao2019dynamic,huang2020lsgcn}}&\multirow{2}{*}{$H^{(i)}=(Z^{(i)}\ast W_1)\odot \sigma(Z^{(i)}\ast W_2)$ 
    \begin{minipage}{1cm}\begin{equation}
        \hspace{5pt}
        \label{eq:GDCC}
    \end{equation}\end{minipage}}\\
    &&\textbf{(GDCC)}&&\\
    
    \cline{2-5}
    
    \rule{0pt}{15pt}&\multirow{2}{*}{{RNN}}&Long Short Term Memory (LSTM)&\cite{lai2018modeling,shih2019temporal}&$H_t^{(i)}=LSTM(Z_t^{(i)}, H_{t-1}^{(i)})$ 
    \begin{minipage}{1cm}\begin{equation}
        \hspace{5pt}
        \label{eq:lstm}
    \end{equation}\end{minipage}
    \cr
    \cline{3-5}
    \rule{0pt}{15pt}&&Gated Recurrent Unit (GRU)&\cite{li2018dcrnn_traffic,chen2020multi,bai2020adaptive}&$H_t^{(i)}=GRU(Z_t^{(i)}, H_{t-1}^{(i)})$ 
    \begin{minipage}{1cm}\begin{equation}
        \hspace{5pt}
        \label{eq:gru}
    \end{equation}\end{minipage}
    \cr
    \cline{2-5}
    
    \rule{0pt}{15pt}&\multirow{2}{*}{\textbf{Attention}}&Transformer&\cite{park2020st,xu2020spatial}&$H^{(i)}=SoftMax(\frac{(Z^{(i)}W_Q)(Z^{(i)}W_K)^\top}{\sqrt{D^{\prime}}})(Z^{(i)}W_V)$ 
    \begin{minipage}{3cm}\begin{equation}
        \hspace{5pt}
        \label{eq:Trans1}
    \end{equation}\end{minipage}
    \cr
    \cline{3-5}
    \rule{0pt}{15pt}&&
    \textbf{Informer (INF-T)}&\cite{haoyietal-informer-2021}
    &$H^{(i)}=SoftMax(\frac{smp(Z^{(i)}W_Q)(Z^{(i)}W_K)^\top}{\sqrt{D^{\prime}}})(Z^{(i)}W_V)$ 
    \begin{minipage}{1cm}\begin{equation}
        \hspace{5pt}
        \label{eq:inform1}
    \end{equation}\end{minipage}
    \cr
    \cline{1-5}
    
        \multirow{3}{*}{\rotatebox{90}{ \emph{S-Operators\phantom{Invis}}}}&\multirow{2}{*}{\textbf{GCN}}&Chebyshev GCN&\cite{yu2018spatio,guo2019attention,diao2019dynamic,fang2019gstnet,huang2020lsgcn}&$H_t=\sum\limits_{k=0}^{K-1}W_kT_k(\Tilde{L})Z_t$ 
        \begin{minipage}{1cm}\begin{equation}
            \hspace{5pt}
            \label{eq:cheb}
        \end{equation}\end{minipage}
        \cr
        \cline{3-5}
    &&\textbf{Diffusion GCN (DGCN)}&\cite{li2018dcrnn_traffic,Wu2019graph,pan2019urban}&$H_t=\sum\limits_{k=0}^{K}(D_O^{-1}A)^{k}Z_tW_1^{k}+(D_I^{-1}A^\top)^{k}Z_tW_2^{k}$ 
    \begin{minipage}{1cm}\begin{equation}
        \hspace{5pt}
        \label{eq:diff}
    \end{equation}\end{minipage}
    \cr
    \cline{2-5}
    
    \rule{0pt}{15pt}&{\multirow{2}{*}{\textbf{Attention}}}&{Transformer}&{\cite{park2020st,xu2020spatial}}&{$H_t=SoftMax(\frac{(Z_tW_Q)(Z_tW_K)^\top}{\sqrt{D^{\prime}}})(Z_tW_V)$} 
    \begin{minipage}{1cm}\begin{equation}
        \hspace{5pt}
        \label{eq:trans2}
    \end{equation}\end{minipage}
    \cr
    \cline{3-5}
    
    \rule{0pt}{15pt}&&{\textbf{Informer} (\textbf{INF-S})}&{None}
    &{$H_t=SoftMax(\frac{smp(Z_tW_Q)(Z_tW_K)^\top}{\sqrt{D^{\prime}}})(Z_tW_V)$} 
    \begin{minipage}{1cm}\begin{equation}
        \hspace{5pt}
        \label{eq:trans2}
    \end{equation}\end{minipage}
    \cr
    
    \bottomrule
    \end{tabular}
    \end{threeparttable}
    \label{table1}
\end{table*}

\noindent \textbf{Reducing the gap between the micro-DAG and the derived ST-block.}
%
%
Due to how we reduce a micro-DAG to an ST-Block, there may be a large gap between the  micro-DAG and the derived ST-Block, which may make the derived ST-block suboptimal. In other words, although we have learned an effective micro-DAG, the derived ST-block may not perform as well as the micro-DAG since the derived ST-Block can be very different from the micro-DAG. 

\begin{figure}[htbp]
\center
\subfigure[Big gap, softmax]{
\begin{minipage}[c]{0.47\linewidth}
\centering
\includegraphics[width=\linewidth]{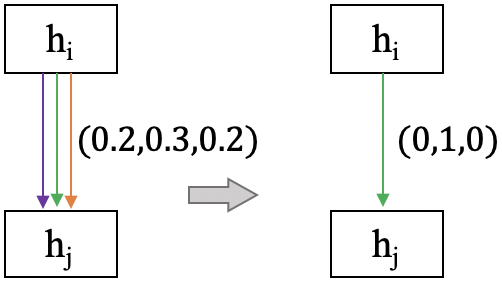}
\label{gap1}
\end{minipage}
}
\subfigure[Small gap, softmax with temperature $\tau$]{
\begin{minipage}[c]{0.47\linewidth}
\centering
\includegraphics[width=\linewidth]{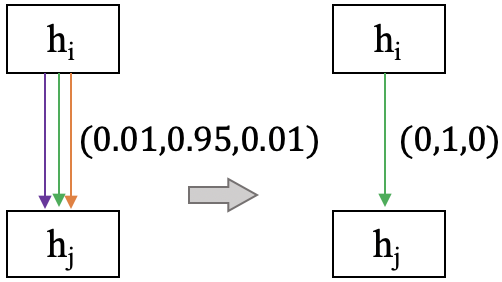}
\label{gap2}
\end{minipage}
}
\caption{Gap between Micro-DAG and ST-Block. 
}
\label{gap}
\end{figure}

Figure~\ref{gap1} shows an example. 
After training the micro-DAG, we get weight vector $\langle 0.2, 0.3, 0.2 \rangle$. To derive the ST-block, we retain the operator with the largest weight, e.g., the second operator. However, the other two operators have relatively high weights as well, meaning that they also contribute significantly to the transformation from $h_i$ to $h_j$. In contrast, in the derived ST-block, only the second operator contributes to the transformation. This represents a big gap, which makes the derived ST-block may not be optimal. 

%
To reduce the discrepancy between the derived ST-block and the extended micro-DAG, we introduce a temperature parameter $\tau$ to the $SoftMax$ function when normalizing parameter $\alpha$, where we replace $\frac{exp(\alpha_o^{(i,j)})}{\sum_{o'\in \mathcal{O}}exp(\alpha_{o'}^{(i,j)})}$ by $\frac{exp(\alpha_o^{(i,j)}/\tau)}{\sum_{o'\in \mathcal{O}}exp(\alpha_{o'}^{(i,j)}/\tau)}$. Thus, $h_j$ is computed as follows. 
$$h_j=\sum\limits_{i<j}\frac{exp(\beta^{(j)}[i])}{\sum_{i<j}exp(\beta^{(j)}[i])}\sum\limits_{o\in \mathcal{O}}\frac{exp(\alpha_o^{i,j}/\tau)}{\sum_{o'\in \mathcal{O}}exp(\alpha_{o'}^{i,j}/\tau)}o(h_i),$$
%
When $\tau \rightarrow 0$, the output of $SoftMax$ is getting closer to a one-hot vector. In this way, for each node pair $(h_i, h_j)$, the operator with largest $\alpha_o^{i,j}$ is dominant, thus making it very close to the derived ST-block that only retains the operator with the largest weight. 
Figure~\ref{gap2} shows an example with a small gap when using the temperature parameter $\tau$. 
In the micro-DAG, the 2nd operator plays a dominant role for the transformation from $h_i$ to $h_j$, and the other two operators contribute only slightly. 
Thus, when ignoring the other two operators in the derived ST-block, the gap is insignificant. 
This ensures that the micro-DAG with optimal architecture parameters $\alpha$ and $\beta$ is able to derive a high-performance ST-block. 

In practice, if we set the temperature $\tau$ to be very small from the beginning, the training process can be unstable. Therefore, we set the initial value of $\tau$ to be relatively large, and perform exponential annealing on $\tau$ to reduce it gradually as training epochs increase. 

\subsubsection{Reducing Operator Set $\mathcal{O}$}
\label{sssec:reduce}
%
%
%
%
%
{Rather than using all S/T operators in the literature, we propose two principles to select a compact set of S/T operators to construct $\mathcal{O}$ with the aim of achieving high search efficiency without compromising accuracy. 
%
%
%
First, selecting S/T operators that capture 
different perspectives is purposeful.
To this end, we categorize existing S/T operators according to their characteristics. 
Second, for each category of S/T operators, we choose the most effective variant. This helps reduce operator set $\mathcal{O}$ without losing promising operators. 
%
}

We categorize commonly used S/T operators for correlated time series forecasting in Table~\ref{table1}. 
Specifically, we categorize the T-operators into three families---the Convolutional Neural Network (CNN) family, the Recurrent Neural Network (RNN) family, and the Attention family; and we categorize the S-operators into two families---the Graph Convolution Network (GCN) family and the Attention family. 

For all equations in Table~\ref{table1}, $Z\in \mathbb{R}^{N\times T\times D}$ denotes the input tensor and $H\in \mathbb{R}^{N\times T^{\prime}\times D^{\prime}}$ denotes the output tensor after applying an S/T operator to $Z$. 
Here, $N$ represents the number of time series or nodes in the graph, $T$ and $T^{\prime}$ represent the number of timestamps, and $D$ and $D^{\prime}$ represent the number of features. 
We use $Z^{(i)}\in \mathbb{R}^{T\times D}, H^{(i)}\in \mathbb{R}^{T\times D^{\prime}}$ to represent the input and output of the $i$-th time series and $Z_{t}\in \mathbb{R}^{N\times D}, H_{t}\in \mathbb{R}^{N\times D^{\prime}}$ to represent the input and output of the $t$-th timestamp. 
We use $A$ to represent the adjacency matrix; $W$ denotes convolution kernels; $W_Q$, $W_K$, and $W_V$ represent projection matrices used in computing attention scores; $D_O$ and $D_I$ represent the diagonal in-degree and out-degree matrices, respectively; and $T_k(\Tilde{L})$ is the Chebyshev polynomial of the adjacency matrix.
Finally, $\ast$ is the convolution operator, $\sigma$ represents the sigmoid function, $smp(\cdot)$ is a sampling function used in Informer, and $\odot$ is the element-wise product. 

\begin{figure}[htb]
  \centering
  \includegraphics[width=0.6\linewidth]{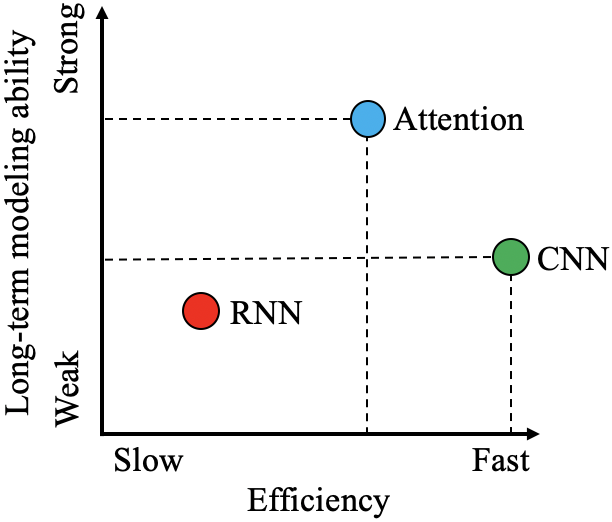}
  \caption{
  Comparison among Different T-operator Families.
  }
  \label{fig:principle}
\end{figure}

\noindent
\textbf{Applying Principle 1: } 
We analyze the different perspectives of different families for T-operators and S-operators, respectively.  
For T-operators, we consider two perspectives---(i) the ability of modeling long-term temporal dependencies and (ii) efficiency. Since short-term temporal dependencies can be relatively easily captured by all families, we do not consider it as a perspective. 
 %
Figure~\ref{fig:principle} shows the CNN, RNN, and Attention families w.r.t. the two perspectives. 

For the CNN family, the core operation is to convolve multiple kernels, e.g., matrices or vectors, with different parts of $Z$ to extract meaningful features.  
The kernel size is often set to be small, which leads to a small receptive field that considers only local features and limits its ability of modeling long-term temporal dependencies. 
To model long-term temporal dependencies, it is possible to stack multiple CNN layers to expand the receptive filed~\cite{Wu2019graph,wu2020connecting,DBLP:conf/cikm/Kieu0GJ18}. 
Since convolutions at different parts of the input tensors are independent, CNNs can be easily parallelized and thus being very efficient.  
 
For the RNN family, the core operation is to compute a hidden state $H_t=f(Z_t, H_{t-1})$ for each timestamp $t$. 
Since $H_t$ is computed based on $H_{t-1}$, 
i.e., the hidden state from the previous timestamp $t-1$, such recursive computations cannot be parallelized. Thus, RNN is  inefficient. 
 To model long term dependencies, many RNN computations need to be done sequentially. This makes RNN suffer from gradient vanishing and exploding problems
 ~\cite{khandelwal2018sharp}, making RNN difficult to capture long-term dependencies. 
Some recent studies have shown that CNN is able to outperform RNN in capturing long-term dependencies~\cite{2016Neural,oord2016conditional,gehring2017convolutional}. This explains why RNN is placed lower than CNN in Figure~\ref{fig:principle}. 
%
%

For the Attention family, the core operation is, for each timestamp, to compute an attention score with each other timestamp. Then, a weighted sum based on the attention scores can be computed. 
%
Since the attention scores are computed w.r.t. all timestamps, this enables the Attention family to be very good at capturing long-term temporal dependencies. 
In addition, since the attention score computations for different timestamps are independent, they can be easily parallelized and thus being efficient. 
However, the CNN family has better efficiency than the Attention family in practice.

To conclude the discussion on the T-operators, 
we disregard the RNN family from our search space because CNN and Attention are more efficient and capture 
long-term temporal dependencies better. We keep both the CNN and Attention families because one is not better than the other one on both perspectives. We provide empirical evidence to justify this design choice in Section~\ref{Sec 5}.

%

We proceed to analyze three different perspectives for the S-operators---(i) whether an adjacency matrix is required,   (ii) the ability of capturing time-varying spatial correlations, 
and (iii) efficiency.

The GCN family relies on an adjacency matrix that indicates the neighboring nodes of each node. The adjacency matrix is often constructed based on meta information, e.g., the distances among the sensors which produce the time series or is learned from data~\cite{Wu2019graph,bai2020adaptive}.  
Given an adjacency matrix, for each node, graph convolution convolves features of neigboring nodes using a learnable kernel such that the features from the neighbors are aggregated. 
%
%
The aggregations per node are independent and thus can be done in parallel. As a result, GCN is efficient. 
%
The spatial correlations between two time series can be different across time, e.g., the correlations on two roads' traffic time series in peak vs. offpeak hours. 
However, the adjacency matrix is often constructed based on distance that does not change across time, this makes GCN fail to capture dynamic spatial correlations. 

The Attention family computes attention scores w.r.t. all other nodes for each node. Thus, it does not require an adjacency matrix that indicates the neighboring relationships. Next, the attention scores can be computed based on 
hidden representations at different timestamps, 
and thus it is able to capture time-varying dependencies. In terms of efficiency, attention scores at different nodes are independent and thus are parallelizable and efficient.



%
Table 2 summarizes the GCN vs. Attention families w.r.t. the perspectives of interest. We observe that the two families complement each other and thus we keep both families in our search space. 

\begin{table}[h]
\small
    \centering
    \caption{Comparison among Different S-operator Families.}
    \begin{tabular}{c|c|cc}
        \hline
        \emph{Perspectives}&\emph{GCN}&\emph{Attention} \cr
        \hline
    Needs predefined adjacency matrix&Yes&No \cr
    \hline
    Captures time-varying spatial correlations&No&Yes \cr
    \hline
    Efficiency&Fastest&Fast \cr
    \hline
    \end{tabular}
    \label{S-operator}
\end{table}

\noindent
\textbf{Applying Principle 2: }
After determining the relevant S/T operator families, we apply the second principle to choose the most effective variant for each family. 
To do so, we consider two scenarios. If there exist studies that compare the different variants in the same experimental setting, we then directly choose the most effective variant. If such studies do not exist, we conduct experiments to identify the most effective variant. 

For the CNN family, we consider 1D Convolution and Gated Dilated Causal Convolution (GDCC). 
The computations of both operators are shown in Equations~\ref{eq:1DC} and ~\ref{eq:GDCC} in Table~\ref{table1}, which clearly indicates 
%
%
%
%
that GDCC is an enhanced version of 1D convolution.  
In addition, a recent paper~\cite{dauphin2017language} has shown strong empirical evidence that GDCC is more effective than 1D convolution. Thus, for the CNN family, we include only GDCC into operator set $\mathcal{O}$. 
%
%

For the temporal Attention family, 
we have two candidates---  Transformer~\cite{vaswani2017attention} and its more efficient variant, Informer~\cite{haoyietal-informer-2021}. 
%
%
Informer improves the attention mechanism in Transformer by only sampling a subset 
of timestamps to calculate the attention score with all the other timestamps, denoted by $smp(\cdot)$
in Equation~\ref{eq:inform1}.  
%
%
In addition, Informer has been shown to be able to also achieve more accurate forecasting than Transformer on time series forecasting tasks~\cite{haoyietal-informer-2021}. 
Thus, we include only Informer, denoted by INF-T as it concerns temporal dependencies, into $\mathcal{O}$.  

For the GCN family, we consider  Chebyshev GCN~\cite{kipf2017semi} and Diffusion GCN~\cite{li2018dcrnn_traffic}. 
%
%
Although the two variants are commonly used in the literature, there is no existing studies compare the two 
variants in a consistent experimental setting for CTS forecasting. We thus
design an experiment to compare them. 
%
This experiment is conducted on two datasets, namely METR-LA and PEMS03 (see the deatils of the two datasets in Section~\ref{ssec:expsetting}). 
The results in Table~\ref{preliminary} show that the diffusion GCN consistently outperforms the Chebyshev GCN. 
Therefore, we include only diffusion GCN into $\mathcal{O}$.
%


\begin{table}[h]
\small
    \centering
    \caption{Comparison of GCN and Attention Variants, MAE.}
    \begin{tabular}{c|c|c||c|c}
        \hline
        &DGCN&Cheby GCN&Informer&Transformer \cr
        \hline
    METR-LA&\textbf{3.33}&{3.42}&\textbf{3.64}&3.65 \cr
    PEMS03&\textbf{18.44}&21.55&23.79&\textbf{23.54} \cr
    \hline
    \end{tabular}
    \label{preliminary}
\end{table}

For the spatial Attention family, 
only Transformer is used in the literature. However, since Informer achieves better accuracy on modeling temporal dependencies, it motivates us to consider it on 
modeling spatial correlations. 
%
We thus conduct an experiment to compare them. 
Table~\ref{preliminary} shows that they have similar accuracy. 
Since Informer is more efficient than Transformer, we include Informer, denoted by INF-S as it concerns spatial correlations, into $\mathcal{O}$.


To summarize, we include GDCC, INF-T, DGCN, and INF-S as the S/T operators in our micro space. In addition, we also include two non-parametric operators, zero and identity. This yields a compact operator set $\mathcal{O}$ with 6 operators.

\subsection{Macro Search Space}
\label{ssec:macro}
We design a macro search space to search for topologies among different ST-blocks. This enables \emph{AutoCTS} to generate ST-backbones with heterogeneous ST-blocks connected by flexible topologies.  

\begin{figure}[htbp]
\center
\subfigure[Marco-DAG]{
\begin{minipage}[c]{0.47\linewidth} 
\centering
\includegraphics[width=\linewidth]{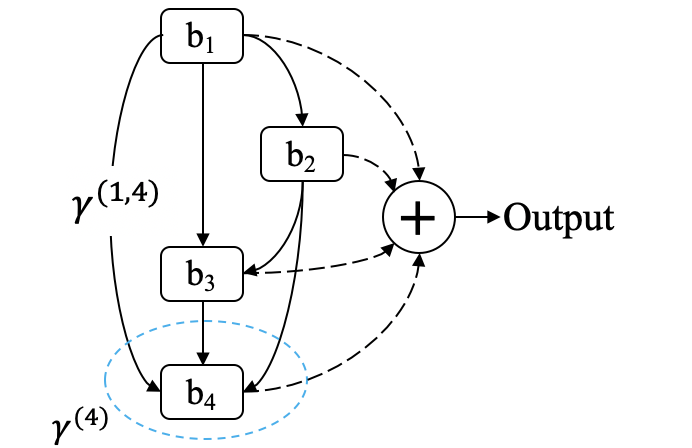}
\label{macro}
\end{minipage}
}
\subfigure[Derived ST-backbone]{
\begin{minipage}[c]{0.47\linewidth}
\centering
\includegraphics[width=\linewidth]{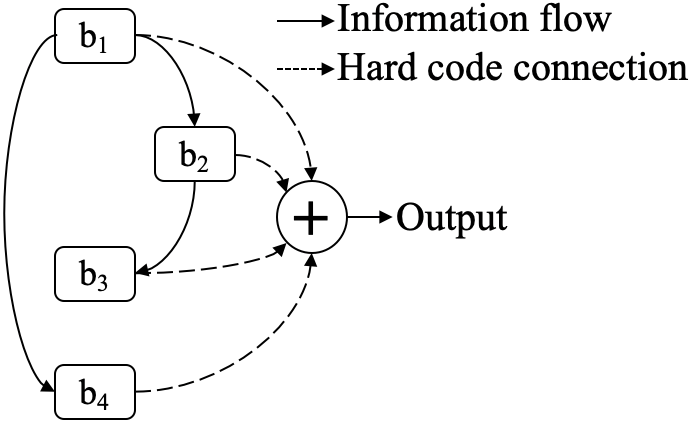}
\label{macro-derive}
\end{minipage}
}
\caption{Marco Search Space. 
}
\label{fig:macro}
\end{figure}

Specifically, we represent the macro search space as a macro-DAG with $B$ nodes, where each node $b_i$, $1\leq i\leq B$, represents an ST-block, and an edge $(b_i, b_j)$ stands for information flow from node $b_i$ to node $b_j$ (see Figure~\ref{macro}). Note that the predecessor of node $b_1$ is the embedding layer.
An information flow from $b_i$ to $b_j$ means that the output representation of ST-block $b_i$ is fed into as the input representation of ST-block $b_j$. This is different from the micro-DAG, where an edge indicates some operators that transform the representations. 

In addition to the information flows which are to be learned, we also have hard code connections from all ST-blocks to the output layer. In other words, no matter which topology is learned to connect the ST-blocks, the outputs of all ST-blocks are merged and fed to the output layer. 

The final, learned ST-backbone is a subgraph of the macro-DAG, where only one incoming edge is retained for each node, i.e., each ST-block (see Figure~\ref{macro-derive}). 

To enable the learning, we introduce the third architecture parameter $\gamma$ to parameterize the information flows among ST-blocks. 
Let $e_{in}^{(j)}$ and $e_{out}^{(j)}$ be the input and output representations of ST-block $b_j$, respectively. We use a scalar-valued parameter $\gamma^{(i,j)}$ 
to represent the weight of edge $(b_i, b_j)$, and calculate $e_{in}^{(j)}$ as the weighted sum of all its predecessors' outputs.  
\begin{align}
e_{in}^{(j)}=\sum\limits_{i<j}\frac{exp(\gamma^{(i,j)})}{\sum_{i<j}exp(\gamma^{(i,j)})}e_{out}^{(i)}
\label{4.3}
\end{align}

At the end of the learning, each ST-block $b_j$ is connected to the precedent $b_i$ with the largest $\gamma^{(i,j)}$. 


To enable heterogeneous ST-blocks, we allow the ST-blocks 
to have different micro architectures. This is achieved by using distinct micro architecture parameters for each ST-block. Specifically, we use architecture parameters $\alpha_i$ and $\beta_i$ to parameterizing the micro search space of ST-block $b_i$. 
The joint search space, which is composed of both the micro and macro search spaces, is parameterized by 
$\Theta=(\{\alpha_i, \beta_i\},  \gamma)$. 




\subsection{Search Strategy}
\label{sec 4.3}


The goal of the architecture search is to learn the architecture parameter $\Theta=(\{\alpha_i, \beta_i\},  \gamma)$ by training the macro-DAG, governed by $\gamma$, and multiple heterogeneous micro-DAGs governed by $\{\alpha_i, \beta_i\}$, in an end-to-end manner. 
%
We design a search strategy to achieve this. 


The learning of \emph{AutoCTS} follows a two-stage strategy---(i) \emph{architecture search} and (ii) \emph{architecture evaluation}. 
In the \emph{architecture search} stage, we run \emph{AutoCTS} on the training set to search for an optimal ST-backbone. 
To do this, we first divide the training data evenly into a pseudo-training data $\mathcal{D}_{train}$ and a pseudo-validation data $\mathcal{D}_{val}$, which are used to train both the architecture parameters $\Theta$ and the network weights $w$, e.g., kernels in CNNs and GCNs, projection matrices in Attentions. 
Specifically, we adopt a bi-level optimization algorithm to optimize $\Theta$ and $w$. 
\begin{align}
\mathop{min}\limits_{\Theta} \quad 
\mathcal{L}_{val}(w^{\ast}, \Theta)
\end{align}
\begin{align}
s.t. \quad w^{\ast}
=\mathop{argmin}\nolimits_w\mathcal{L}_{train}(w,\Theta),
\end{align}
where $\mathcal{L}_{train}$ and $\mathcal{L}_{val}$ denote the losses (e.g., mean absolute error or mean squared error) on the pseudo-training and pseudo-validation data, respectively. 
We employ first-order approximation to speed-up the architecture search~\cite{liu2018darts}. The detailed training process is shown in {Algorithm~\ref{algorithm}}, where $\eta$ and $\xi$ are the learning rates for the two optimizers for $\Theta$ and $w$, respectively. 

\begin{algorithm}[htb]
\caption{Joint Search Algorithm}
\hspace*{0.02in} {\bf Input:} 
Correlated time series $\mathcal{X} \in R^{N\times T\times F}$, Adjacency matrix $G$; 
\label{algorithm}
\begin{algorithmic}[1]
\State Randomly initialize $\Theta=(\{\alpha_i, \beta_i\},  \gamma)$ and $w$. Split training data into pseudo train data $\mathcal{D}_{train}$ and pseudo validation data $\mathcal{D}_{val}$.
\State {\bf While} Not exceeding the largest epoch {\bf do}
\State \hspace{0.2in} Sample a mini-batch from $\mathcal{D}_{val}.$
\State \hspace{0.2in} Update $\Theta$ with $\Theta=\Theta-\eta \bigtriangledown_\Theta\mathcal{L}_{val}(w, \Theta)$ 
\State \hspace{0.2in} Sample a mini-batch from $\mathcal{D}_{train}.$
\State \hspace{0.2in} Update $w$ with $w=w-\xi \bigtriangledown_w \mathcal{L}_{train}(w,\Theta)$ 
.
\State {\bf return} ST-backbone w.r.t. the learned 
$\Theta$.
\end{algorithmic}
\end{algorithm}


In the \emph{architecture evaluation} stage, we only keep the architecture parameters $\Theta$ but discard the learned network weights $w$ from the architecture search stage. This means that we only keep the learned neural architecture of the learned ST-backbone. Instead, we train the forecasting model with the learned ST-backbone from scratch on the original training and validation sets to obtain new network weights $w^\prime$. Finally, we report the accuracy of the forecasting model with $w^\prime$ on the testing set.




\section{Experiments}
\label{Sec 5}
We evaluate \emph{AutoCTS} on both single- and multi-step time series forecasting tasks using eight correlated time series datasets from different domains to justify our design choices.
%

\subsection{Experimental Settings}
\label{ssec:expsetting}
\subsubsection{Datasets}
To enable fair comparisons with existing studies and to facilitate reproducibility, we employ eight commonly used benchmark datasets for correlated time series forecasting, including six datasets for multi-step forecasting~\cite{yu2018spatio,guo2019attention,li2018dcrnn_traffic,Wu2019graph,bai2020adaptive,song2020spatial} and two datasets for single-step forecasting~\cite{lai2018modeling,shih2019temporal}.

\noindent
\textbf{Multi-step forecasting: }

\begin{itemize}[leftmargin=*]
    \item METR-LA and PEMS-BAY: Both datasets are traffic speed time series datasets, released by Li et al.~\cite{li2018dcrnn_traffic}. The two datasets are collected from highways in the Los Angeles County and the Bay area, respectively. 
    %
    %
    \item PEMS03, PEMS04, PEMS07 and PEMS08: All datasets are traffic flow time series collected from the Caltrans Performance Measurement System (PeMS), which are released by Song et al.~\cite{song2020spatial}. 
\end{itemize}
Table~\ref{table3} summarizes the statistics of the six datasets. This includes $N$, the number of time seriesor nodes, and $T$, the total number of timestamps. We adopt the same 
train-validation-test splits as in the original papers~\cite{li2018dcrnn_traffic, song2020spatial}, as shown in the ``Split Ratio'' column in Table~\ref{table3}. 
All the time series in the six dataset have a record every 5 minutes, and thus there are 12 records per hour. 
Following existing literature~\cite{li2018dcrnn_traffic,Wu2019graph,wu2020connecting}, we consider a multi-step forecasting setting where we use the recent one hour in the history (i.e., input=12 timestamps) to forecast the records in the next hour (i.e., output=12 timestamps). 
For each dataset, a graph is constructed where each node represents a sensor that generates a time series. The adjacency matrix represents the road network distances among the sensors \cite{yu2018spatio,li2018dcrnn_traffic,Wu2019graph,song2020spatial}.

\begin{table}[h]
\small
    \centering
    \caption{Datasets.}
    \begin{tabular}{l l l l l l}
        \hline
         Dataset & $N$ & $T$ & Split Ratio & Input & Output\\
        \hline\hline
         METR-LA & 207 & 34,272 & 7:1:2 & 12 & 12\\
         PEMS-BAY & 325 & 52,116 & 7:1:2 & 12 & 12\\
         PEMS03 & 358 & 26,208 & 6:2:2 & 12 & 12\\
         PEMS04 & 307 & 16,992 & 6:2:2 & 12 & 12\\
         PEMS07 & 883 & 28,224 & 6:2:2 & 12 & 12\\
         PEMS08 & 170 & 17,856 & 6:2:2 & 12 & 12\\
        \hline
         Solar-energy & 137 & 52,560 & 6:2:2 & 168 & 1\\
         Electricity & 321 & 26,304 & 6:2:2 & 168 & 1\\
        \hline
    \end{tabular}
    \label{table3}
\end{table}

To enable direct and fair comparisons with existing studies~\cite{li2018dcrnn_traffic,Wu2019graph,wu2020connecting,song2020spatial,bai2020adaptive}, 
for METR-LA and PEMS-BAY, we report accuracy of the forecasts on the 3rd, 6th, and 12th timestamps, corresponding to the next 15-min, 30-min, and 60-min, respectively; for PEMS03, PEMS04, PEMS07, and PEMS08, we report the average accuracy over all 12 future timestamps.

\noindent
\textbf{Single-step forecasting: }
%

\begin{itemize}[leftmargin=*]
    \item Solar-Energy: The solar power production records collected from 137 PV plants in the Alabama State, released by Lai et al. ~\cite{lai2018modeling}
    \item Electricity: The electricity consumption records collected from 321 clients, released by Lai et al. ~\cite{lai2018modeling}.
\end{itemize}
The statistics of the two datasets are also summarized in Table \ref{table3}.
We also use the same 
train-validation-test splits as the original paper~\cite{lai2018modeling}. Similar to the multi-step forecasting, we consider a well-known single-step forecasting setup to enable fair comparions with existing studies. Specifically, we use the historical 168 timestamps (i.e., input=168 timestamps) to predict the value in a single future timestamp (i.e., output=1 timestamp). The single future timestamp is either 3 or 24. 
There is no predefined adjacency matrix for Solar-Energy and Electricity datasets. 

\subsubsection{Evaluation Metrics}
Following the evaluation methodology in previous studies~\cite{li2018dcrnn_traffic,Wu2019graph,lai2018modeling,wu2020connecting}, we use mean absolute error (MAE), root mean squared error (RMSE), mean absolute percentage error (MAPE) to evaluate the accuracy of multi-step forecasting, and use Root Relative Squared Error (RRSE) and Empirical Correlation Coefficient (CORR) to measure the accuracy of single-step forecasting. For MAE, RMSE, MAPE, and RRSE, lower values indicate higher accuracy, while larger CORR values indicate higher accuracy.

\subsubsection{Baselines}
We compare \emph{AutoCTS} with eight methods, including seven methods that are manually designed by human experts and one automated approach. 
The implementations of the baselines are based on the public-available code released by their authors. 
\begin{itemize}[leftmargin=*]
    \item DCRNN: Diffusion convolutional recurrent neural network uses diffusion GCN
    with GRU to build ST-blocks, and employs an encoder-decoder architecture for multi-step forecasting~\cite{li2018dcrnn_traffic}.
    \item[$\bullet$] STGCN: Spatio-temporal graph convolutional network adopts a Chebyshev GCN and a gated 1D convolution to build ST-blocks ~\cite{yu2018spatio}.
    \item[$\bullet$] Graph WaveNet: 
    It employs diffusion GCN and GDCC to build ST-blocks ~\cite{Wu2019graph}.
    \item[$\bullet$] AGCRN: Adaptive graph convolutional recurrent network combines enhanced Chebyshev GCN and GRU to build  ST-blocks~\cite{bai2020adaptive}. 
    \item[$\bullet$] LSTNet: A long- and short-term time-series network, which combines 1D convolution and GRU to extract short-term and long-term temporal dependencies~\cite{lai2018modeling}.
    \item[$\bullet$] TPA-LSTM: An attention based recurrent neural network~\cite{shih2019temporal}.
    \item[$\bullet$] MTGNN: A multivariate time series forecasting model with graph neural networks, which utilizes a spatial-based GCN and GDCC to construct ST-blocks~\cite{wu2020connecting}.
    \item[$\bullet$] AutoSTG: A NAS based method for automated spatio-temporal graph prediction, which uses only diffusion GCN and 1D convolution as the S/T operators in the search space for only ST-blocks but not ST-backbones, and employs meta learning to learn the weights for the diffusion GCN and 1D convolution~\cite{pan2019urban}.
\end{itemize}

\subsubsection{Implementation Details}
All the model training experiments are conducted on Nvidia Quadro RTX 8000 GPUs. The source code is available at \url{\vldbavailabilityurl}.

\noindent \textbf{Architecture Search. }
Following Liu et al.~\cite{liu2018darts}, we use the ReLU-operator-BN order for all parametric operators to improve the training stability.
We vary the number of nodes in the micro-DAG $M$ among 3, \textbf{5}, and 7, and vary the number of nodes in the macro-DAG $B$ among 2, \textbf{4}, 6, with default values shown in bold{, for all datasets.} 
We adopt Adam~\cite{kingma2014adam} as the optimizer for both the architecture parameters $\Theta$ and the network weights $w$. For $\Theta$, we set the initial learning rate to $3\times 10^{-4}$, the momentum to $(0.5, 0.999)$, and the weight decay to $10^{-3}$. For $w$, we set the initial learning rate to $10^{-3}$, and the weight decay to $10^{-4}$.
%
We adopt partial channels~\cite{xu2019pc} to improve the memory efficiency, where we select $1/4$ features during training. 
%
{For all datasets, we set the initial temperature $\tau$ to $5.0$ and use exponential annealing with a multiplicative factor of $0.9$ to gradually reduce it as training evolves until it reaches $0.001$. }

\subsection{Experimental Results}
\begin{table*}[h]
\small
    \centering
    \caption{Accuracy of Multi-step Forecasting, METR-LA and PEMS-BAY.}
    \begin{threeparttable} 
    \begin{tabular}{ccccc|ccc|ccc}
        \hline
        \toprule  
        \multirow{2}{*}{Data}&
        \multirow{2}{*}{Models}&
        
        \multicolumn{3}{c}{15 min}&
        \multicolumn{3}{c}{30 min}&
        \multicolumn{3}{c}{60 min} \cr  
    & &MAE&RMSE&MAPE&MAE&RMSE&MAPE&MAE&RMSE&MAPE\cr  
    \midrule  
    \multirow{7}{*}{\rotatebox{90}{METR-LA}}& DCRNN& 2.77&5.38&7.30\%&3.15&6.45&8.80\%&3.60&7.60&10.50\% \cr
    & STGCN& 2.88&5.74&7.62\%&3.47&7.24&9.57\%&4.59&9.40&12.70\% \cr
    & Graph WaveNet& \underline{2.69}&\underline{5.15}&6.90\%&3.07&6.22&8.37\%&3.53&7.37&10.01\% \cr
    & AGCRN& 2.83&5.45&{7.56}\%&{3.20}&{6.55}&{8.79}\%&3.58&{7.41}&{10.13}\% \cr
    & MTGNN& \underline{2.69}&5.18&\underline{6.86}\%&\textbf{3.05}&\underline{6.17}&\underline{8.19}\%&\underline{3.49}&\underline{7.23}&\underline{9.87}\% \cr
    & AutoSTG& 2.70&5.16&6.91\%&\underline{3.06}&\underline{6.17}&8.30\%&\textbf{3.47}&{7.27}&\underline{9.87}\% \cr
    & \emph{AutoCTS}& \textbf{2.67}&\textbf{5.11}&\textbf{6.80}\%&\textbf{3.05}&\textbf{6.11}&\textbf{8.15}\%&\textbf{3.47}&\textbf{7.14}&\textbf{9.81}\% \cr
    \hline
    \multirow{7}{*}{\rotatebox{90}{PEMS-BAY}}& DCRNN& 1.38&2.95&2.90\%&1.74&3.97&3.90\%&2.07&4.74&4.90\% \cr
    & STGCN& 1.36&2.96&2.90\%&1.81&4.27&4.17\%&2.49&5.69&5.79\% \cr
    & Graph WaveNet& \textbf{1.30}&\underline{2.74}&\underline{2.73}\%&\underline{1.63}&3.70&{3.67}\%&1.95&4.52&4.63\% \cr
    & AGCRN& 1.35&2.83&2.87\%&{1.69}&{3.81}&{3.84}\%&1.96&{4.52}&{4.67}\% \cr
    & MTGNN&1.32&2.79&2.77\%&1.65&3.74&3.69\%&1.94&4.49&{4.53}\% \cr
    & AutoSTG& \underline{1.31}&2.76&\underline{2.73}\%&\underline{1.63}&\underline{3.67}&\underline{3.63}\%&\underline{1.92}&\underline{4.38}&\underline{4.43}\% \cr
    & \emph{AutoCTS}& \textbf{1.30}&\textbf{2.71}&\textbf{2.69}\%&\textbf{1.61}&\textbf{3.62}&\textbf{3.55}\%&\textbf{1.89}&\textbf{4.32}&\textbf{4.36}\% \cr
    \bottomrule
    \end{tabular}
    \end{threeparttable}
    \label{table4}
\end{table*}

\begin{table*}[!htb]
    \centering
    \begin{minipage}{.72\textwidth}
        \centering
         \caption{Accuracy of Multi-step Forecasting, PEMS03, PEMS04, PEMS07, and PEMS08.}
    \begin{tabular}{c|c|ccccc|c}
        \hline
        Data&Metric&DCRNN&STGCN&Graph WaveNet&AGCRN&MTGNN&\emph{AutoCTS} \cr
        \hline
    \multirow{3}{*}{PEMS03}&MAE&18.18&17.49&\underline{14.82}&15.89&15.10&\textbf{14.71} \cr
     &RMSE&30.31&30.12&\underline{25.24}&28.12&25.93&\textbf{24.54} \cr
     &MAPE&18.91\%&17.15\%&16.16\%&\underline{15.38}\%&{15.67}\%&\textbf{14.39}\% \cr
    \hline
    \multirow{3}{*}{PEMS04}&MAE&24.70&22.70&\underline{19.16}&19.83&19.32&\textbf{19.13} \cr
     &RMSE&38.12&35.55&\underline{30.46}&32.26&31.57&\textbf{30.44} \cr
     &MAPE&17.12\%&14.59\%&13.26\%&\underline{12.97}\%&13.52\%&\textbf{12.89}\% \cr
    \hline
    \multirow{3}{*}{PEMS07}&MAE&25.30&25.38&21.54&\underline{21.31}&22.07&\textbf{20.93} \cr
     &RMSE&38.58&38.78&\underline{34.23}&35.06&35.80&\textbf{33.69} \cr
     &MAPE&11.66\%&11.08\%&9.22\%&\underline{9.13}\%&9.21\%&\textbf{8.90}\% \cr
    \hline
    \multirow{3}{*}{PEMS08}&MAE&17.86&18.02&\underline{15.13}&15.95&15.71&\textbf{14.82} \cr
     &RMSE&27.83&27.83&\underline{24.07}&25.22&24.62&\textbf{23.64} \cr
     &MAPE&11.45\%&11.40\%&10.10\%&{10.09}\%&\underline{10.03}\%&\textbf{9.51}\% \cr
    \bottomrule
    \end{tabular}
    \label{table5}
        
    \end{minipage}%
    \begin{minipage}{0.28\textwidth}
        \centering
     {
     \caption{Search time (GPU hours) and memory (MB).}
    \begin{tabular}{ccc}
        \toprule  
        \multirow{2}{*}{DataSet}&
        {Search}&
        \multirow{2}{*}{Memory}\\
        &Time& \\
    \midrule  
    METR-LA&21.43&20,109 \cr
    PEMS-BAY&52.60&32,117 \cr
    PEMS03&{25.95}&34,401 \cr
    PEMS04&{14.64}&30,681 \cr
    PEMS07&{61.74}&33,057 \cr
    PEMS08&{12.34}&16,521 \cr
    Solar-Energy&{163.21}&30,977 \cr
    Electricity&{145.68}&36,339 \cr
    \bottomrule
    \end{tabular}
    \label{memory}
    }
    \end{minipage}
\end{table*}

\subsubsection{Multi-step Forecasting Accuracy}
Tables~\ref{table4} and~\ref{table5} present the overall accuracy of \emph{AutoCTS} and the baselines on multi-step forecasting datasets. 
We use bold to highlight the best accuracy and underline the second best accuracy.
%
Since AutoSTG relies on additional information on the road network to enable meta learning based weight generation, and such information is unavailable on the four PEMS datasets, AutoSTG is thus unable to work on the four PEMS datasets. 

Key observations are as follows. 
First, \emph{AutoCTS} outperforms all manually designed models 
on all multi-step forecasting tasks, demonstrating that \emph{AutoCTS} is able to produce very competitive ST-blocks and ST-backbones that outperform human designed models. 

Second, when comparing to the other automated approach AutoSTG, although AutoSTG includes additional features, such as GPS coordinates of sensors, to enhance its forecasting accuracy, it is still inferior to \emph{AutoCTS} except for the MAE at the 60-min timestamp on METR-LA. This is due to (i) AutoSTG only include diffusion GCN and 1D convolution to construct the search space, while we follow the proposed two principles to select a compact yet complementary S/T operators. 
(ii) AutoSTG only searches for the micro architecture of a single ST-block, and then stacks the ST-blocks to build the forecasting model. In contrast, we jointly search for both the micro architecture of ST-blocks and the macro architecture of the ST-backbone. 

Third, \emph{AutoCTS} outperforms AGCRN and DCRNN, which both employ GRU to model temporal dependencies. This justifies our design choices that disregard the RNN family in the micro search space. 

Fourth, there does not exist a single manually-designed model that consistently outperforms 
other manually-designed models. For example, MTGNN outperforms Graph WaveNet on METR-LA, but is outperformed by Graph WaveNet on PEMS04. 
This suggests that the optimal neural architectures for different datasets may be different, which implies that it is beneficial to be able to 
automatically identify forecasting models with unique architectures for different datasets, which is what \emph{AutoCTS} is able to offer.

\subsubsection{Single-step Forecasting Accuracy}
Table \ref{table6} shows the experimental results on the two single-step forecasting datasets. 
We observe that: (1) 
\emph{AutoCTS} and MTGNN outperform LSTNet and TPA-LSTM. This is because  LSTNet and TPA-LSTM do not explicitly model the correlations among different time series. In contrast, both \emph{AutoCTS} and MTGNN simultaneously model the temporal and spatial dependencies. {\emph{AutoCTS} does not outperform MTGNN much for single-step forecasting when compared to multi-step forecasting. This suggests that MTGNN is already a very effective model that behaves similarly to the optimal model identified from the search space of \emph{AutoCTS} for single-step forecasting, but is less effective for multi-step forecasting. This further justifies the needs for automated solutions for identifying specific, optimal models for different forecasting tasks. }
(2) \emph{AutoCTS} achieves the best accuracy on both short-term (see Table~\ref{table4} and Table~\ref{table5}) and long-term (see Table~\ref{table6}) datasets. This is because our search space contains GDCC and INF-T, which are good at modeling both short- and long-term dependencies, respectively, which enables \emph{AutoCTS} to generate high-performance models in both cases. 


\begin{table}[htbp]
\small
    \centering
    \caption{Accuracy of Single-step Forecasting.}
    \begin{tabular}{c|c|c|c|c|c}
        \toprule  
        \multicolumn{2}{c|}{Data}&
        \multicolumn{2}{c|}{Solar-Energy}&        \multicolumn{2}{c}{Electricity} \cr
    \hline
    Models&Metric&3&24&3&24 \cr
    \midrule 
    \multirow{2}{*}{LSTNet}&RRSE&0.1843&0.4643&0.0864&0.1007 \cr
     &CORR&0.9843&0.8870&0.9283&0.9119 \cr
    \hline
    \multirow{2}{*}{TPA-LSTM}&RRSE&0.1803&0.4389&0.0823&0.1006 \cr
     &CORR&0.9850&\underline{0.9081}&0.9439&0.9133 \cr
    \hline
    \multirow{2}{*}{MTGNN}&RRSE&\underline{0.1778}&\underline{0.4270}&\underline{0.0745}&\underline{0.0953} \cr
     &CORR&\underline{0.9852}&0.9031&\underline{0.9474}&\underline{0.9234} \cr
    \hline
    \multirow{2}{*}{\emph{AutoCTS}}&RRSE&\textbf{0.1750}&\textbf{0.4143}&\textbf{0.0743}&\textbf{0.0947} \cr
     &CORR&\textbf{0.9855}&\textbf{0.9085}&\textbf{0.9477}&\textbf{0.9239} \cr
    \bottomrule
    \end{tabular}
    \label{table6}
\end{table}

\subsubsection{Ablation Studies}
\label{ablation}
We conduct ablation studies to justify the design choices used in 
\emph{AutoCTS}. 
\begin{table*}[!htbp]
\small
    \centering
    \caption{Ablation Studies, METR-LA}
    \vspace{-10pt}
    \begin{tabular}{cccc|ccc|ccc|c}
        \toprule  
        \multirow{2}{*}{Models}&
        \multicolumn{3}{c}{15 min}&
        \multicolumn{3}{c}{30 min}&
        \multicolumn{3}{c}{60 min}&
        \multirow{2}{*}{GPU hours} \cr
    &MAE&RMSE&MAPE&MAE&RMSE&MAPE&MAE&RMSE&MAPE\cr  
    \midrule  
    \emph{AutoCTS}&\textbf{2.67}&{5.11}&{6.80}\%&\textbf{3.05}&{6.11}&{8.15}\%&\textbf{3.47}&\textbf{7.14}&{9.81}\%&21.43 \cr
    w/o design principles&2.72&5.20&6.97\%&3.12&6.27&8.42\%&3.57&7.34&9.98\%&96.79 \cr
    w/o temperature&2.75&5.21&6.89\%&3.13&6.19&8.23\%&3.55&7.20&9.84\%&21.42 \cr
    w/o macro search&2.68&\textbf{5.07}&\textbf{6.76}\%&3.07&\textbf{6.09}&\textbf{8.08}\%&3.56&7.22&\textbf{9.75}\%&{21.37} \cr
    macro only&2.73&5.21&7.02\%&3.07&6.20&8.34\%&3.51&7.29&10.02\%&\textbf{10.97}
    \cr
    \bottomrule
    \end{tabular}
    \label{table_la_ablation}
    \vspace{-5pt}
\end{table*}

\begin{table*}[!htbp]
\small
    \centering
    \caption{Ablation Studies, PEMS-BAY}
    \vspace{-10pt}
    \begin{tabular}{cccc|ccc|ccc|c}
        \toprule  
        \multirow{2}{*}{Models}&
        \multicolumn{3}{c}{15 min}&
        \multicolumn{3}{c}{30 min}&
        \multicolumn{3}{c}{60 min}&
        \multirow{2}{*}{GPU hours} \cr
    &MAE&RMSE&MAPE&MAE&RMSE&MAPE&MAE&RMSE&MAPE\cr  
    \midrule  
    \emph{AutoCTS}&\textbf{1.30}&\textbf{2.71}&\textbf{2.69}\%&\textbf{1.61}&\textbf{3.62}&\textbf{3.55}\%&\textbf{1.89}&\textbf{4.32}&\textbf{4.36}\%&52.60 \cr
    w/o design principles&\textbf{1.30}&2.72&2.71\%&1.62&3.64&3.62\%&1.92&4.38&4.52\%&223.56 \cr
    w/o temperature&1.31&2.72&2.72\%&1.63&3.65&3.68\%&1.93&4.42&4.54\%&52.63 \cr
    w/o macro search&1.31&2.74&\textbf{2.69}\%&1.62&3.66&\textbf{3.55}\%&\textbf{1.89}&4.35&4.40\%&{52.31} \cr
    macro only&1.34&2.77&2.83\%&1.65&3.68&3.74\%&1.95&4.42&4.66\%&\textbf{24.08}
    \cr
    \bottomrule
    \end{tabular}
    \label{table_bay_ablation}
    \vspace{-5pt}
\end{table*}

\begin{table*}[!htb]
    \centering
    \begin{minipage}{.5\textwidth}
        \centering
         \caption{Ablation Studies, PEMS03}
    \vspace{-10pt}
    \begin{tabular}{ccccc}
        \toprule  
        \multirow{2}{*}{Models}&
        \multicolumn{3}{c}{Metrics}&
        \multirow{2}{*}{GPU hours} \cr
    &MAE&RMSE&MAPE\cr  
    \midrule  
    \emph{AutoCTS}& \textbf{14.71}&\textbf{24.54}&\textbf{14.39}\%&25.95 \cr
    w/o design principles& 15.66&25.51&15.28\%&126.25 \cr
    w/o temperature&14.87&24.93&14.64\%&25.97 \cr
    w/o macro search&  15.07&25.22&14.84\%&{25.89} \cr
    macro only&15.83&26.12&15.77\%&\textbf{15.90}
    \cr
    \bottomrule
    \end{tabular}
    \label{table_03_ablation}
        
    \end{minipage}%
    \begin{minipage}{0.5\textwidth}
        \centering
     \caption{Ablation Studies, PEMS04}
    \vspace{-10pt}
    \begin{tabular}{ccccc}
        \toprule  
        \multirow{2}{*}{Models}&
        \multicolumn{3}{c}{Metrics}&
        \multirow{2}{*}{GPU hours} \cr
    &MAE&RMSE&MAPE\cr  
    \midrule  
    \emph{AutoCTS}& \textbf{19.13}&\textbf{30.44}&\textbf{12.89}\%&14.64 \cr
    w/o design principles&19.33&30.64&12.98\%&48.40 \cr
    w/o temperature&19.23&30.59&13.03\%&14.64 \cr
    w/o macro search&19.59&30.86&13.07\%&{14.60} \cr
    macro only&20.14&31.41&13.97\%&\textbf{8.27}
    \cr
    \bottomrule
    \end{tabular}
    \label{table_04_ablation}
    \end{minipage}
\end{table*}

\begin{table*}[!htb]
    \centering
    \begin{minipage}{.5\textwidth}
        \centering
         \caption{Ablation Studies, PEMS07}
    \vspace{-10pt}
    \begin{tabular}{ccccc}
        \toprule  
        \multirow{2}{*}{Models}&
        \multicolumn{3}{c}{Metrics}&
        \multirow{2}{*}{GPU hours} \cr
    &MAE&RMSE&MAPE\cr  
    \midrule  
    \emph{AutoCTS}& \textbf{20.93}&\textbf{33.69}&\textbf{8.90}\%&61.74 \cr
    w/o design principles& 21.03&33.78&9.01\%&247.67 \cr
    w/o temperature&21.16&34.12&9.22\%&61.71 \cr
    w/o macro search&21.23&34.02&9.20\%&{61.68} \cr
    macro only&22.91&36.02&10.16\%&\textbf{31.64}
    \cr
    \bottomrule
    \end{tabular}
    \label{table_07_ablation}
        
    \end{minipage}%
    \begin{minipage}{0.5\textwidth}
        \centering
     \caption{Ablation Studies, PEMS08}
    \vspace{-10pt}
    \begin{tabular}{ccccc}
        \toprule  
        \multirow{2}{*}{Models}&
        \multicolumn{3}{c}{Metrics}&
        \multirow{2}{*}{GPU hours} \cr
    &MAE&RMSE&MAPE\cr  
    \midrule  
    \emph{AutoCTS}& \textbf{14.82}&\textbf{23.64}&\textbf{9.51}\%&12.34 \cr
    w/o design principles&15.48&24.21&10.33\%&46.13 \cr
    w/o temperature&15.03&23.69&9.69\%&12.34 \cr
    w/o macro search&15.24&24.17&9.91\%&{12.32} \cr
    macro only&16.03&24.91&10.92\%&\textbf{6.78}
    \cr
    \bottomrule
    \end{tabular}
    \label{table_08_ablation}
    \end{minipage}
\end{table*}

\begin{table*}[!htbp]
\small
    \centering
    \caption{Ablation Studies, Solar-Energy}
    \vspace{-10pt}
    \begin{tabular}{ccc|cc|c}
        \toprule  
        \multirow{2}{*}{Models}&
        \multicolumn{2}{c}{3}&
        \multicolumn{2}{c}{24}&
        \multirow{2}{*}{GPU hours} \cr
    &RRSE&CORR&RRSE&CORR\cr  
    \midrule  
    \emph{AutoCTS}& {0.1750}&{0.9855}&{0.4143}&\textbf{0.9085}&{163.21} \cr
    w/o design principles&0.1783&0.9846&0.4159&0.9062&559.09 \cr
    w/o temperature&\textbf{0.1658}&\textbf{0.9875}&\textbf{0.4109}&0.9084&163.37 \cr
    w/o macro search&0.1763&0.9847&0.4158&{0.9034}&162.70 \cr
    macro only&0.1787&0.9832&0.4183&0.9029&\textbf{98.31}
    \cr
    \bottomrule
    \end{tabular}
    \label{table_solar_ablation}
\end{table*}

\begin{table*}[!htbp]
\small
    \centering
    \caption{Ablation Studies, Electricity}
    \vspace{-10pt}
    \begin{tabular}{ccc|cc|c}
        \toprule  
        \multirow{2}{*}{Models}&
        \multicolumn{2}{c}{3}&
        \multicolumn{2}{c}{24}&
        \multirow{2}{*}{GPU hours} \cr
    &RRSE&CORR&RRSE&CORR\cr  
    \midrule  
    \emph{AutoCTS}& \textbf{0.0743}&\textbf{0.9477}&\textbf{0.0947}&\textbf{0.9239}&145.68 \cr
    w/o design principles&0.0753&0.9464&0.0952&0.9227&494.15 \cr
    w/o temperature&0.0753&0.9457&0.0958&{0.9231}&145.73 \cr
    w/o macro search&0.0780&0.9441&0.0963&{0.9206}&145.62 \cr
    macro only&0.0832&0.9230&0.0950&0.9168&\textbf{85.19}
    \cr
    \bottomrule
    \end{tabular}
    \label{table_elec_ablation}
\end{table*}

\begin{table*}[htbp]
\small
    \centering
    \caption{Impact of $M$, METR-LA.}
    \vspace{-10pt}
    \begin{tabular}{c|ccccccccc}
        \toprule  
        \multirow{2}{*}{M}&
        \multicolumn{3}{c}{15 min}&
        \multicolumn{3}{c}{30 min}&
        \multicolumn{3}{c}{60 min} \cr
    &MAE&RMSE&MAPE&MAE&RMSE&MAPE&MAE&RMSE&MAPE\cr 
    \hline
    3&2.71&5.14&6.91\%&3.08&6.16&8.41\%&3.54&7.26&10.08\% \cr
    5&\textbf{2.67}&\textbf{5.11}&\textbf{6.80}\%&\textbf{3.05}&\textbf{6.11}&\textbf{8.15}\%&\textbf{3.47}&\textbf{7.14}&\textbf{9.81}\% \cr
    7&2.73&5.22&7.07\%&3.11&6.25&8.38\%&3.56&7.37&10.21\% \cr
    \bottomrule
    \end{tabular}
    \label{hyper_laM}
\end{table*}

\begin{table*}[htbp]
\small
    \centering
    \caption{Impact of $B$, METR-LA.}
    \vspace{-10pt}
    \begin{tabular}{c|ccccccccc}
        \toprule  
        \multirow{2}{*}{B}&
        \multicolumn{3}{c}{15 min}&
        \multicolumn{3}{c}{30 min}&
        \multicolumn{3}{c}{60 min} \cr
    &MAE&RMSE&MAPE&MAE&RMSE&MAPE&MAE&RMSE&MAPE\cr 
    \hline
    2&{2.70}&{5.11}&{6.96}\%&{3.07}&\textbf{6.07}&{8.42}\%&{3.50}&\textbf{7.09}&{10.15}\% \cr
    4&\textbf{2.67}&{5.11}&\textbf{6.80}\%&{3.05}&{6.11}&\textbf{8.15}\%&\textbf{3.47}&{7.14}&\textbf{9.81}\% \cr
    6&2.68&\textbf{5.09}&6.88\%&\textbf{3.04}&\textbf{6.07}&8.35\%&3.50&7.18&10.18\% \cr
    \bottomrule
    \end{tabular}
    \label{hyper_laB}
\end{table*}

\begin{table*}[htbp]
\small
    \centering
    \caption{Impact of $M$, PEMS-BAY.}
    \vspace{-10pt}
    \begin{tabular}{c|ccccccccc}
        \toprule  
        \multirow{2}{*}{M}&
        \multicolumn{3}{c}{15 min}&
        \multicolumn{3}{c}{30 min}&
        \multicolumn{3}{c}{60 min} \cr
    &MAE&RMSE&MAPE&MAE&RMSE&MAPE&MAE&RMSE&MAPE\cr 
    \hline
    3&\textbf{1.30}&\textbf{2.70}&{2.71}\%&1.62&3.65&3.65\%&1.91&4.38&4.55\% \cr
    5&\textbf{1.30}&{2.71}&\textbf{2.69}\%&\textbf{1.61}&\textbf{3.62}&\textbf{3.55}\%&\textbf{1.89}&\textbf{4.32}&\textbf{4.36}\% \cr
    7&1.32&2.73&2.76\%&1.64&3.72&3.69\%&1.92&4.48&4.60\% \cr
    \bottomrule
    \end{tabular}
    \label{hyper_bayM}
\end{table*}

\begin{table*}[htbp]
\small
    \centering
    \caption{Impact of $B$, PEMS-BAY.}
    \vspace{-10pt}
    \begin{tabular}{c|ccccccccc}
        \toprule  
        \multirow{2}{*}{B}&
        \multicolumn{3}{c}{15 min}&
        \multicolumn{3}{c}{30 min}&
        \multicolumn{3}{c}{60 min} \cr
    &MAE&RMSE&MAPE&MAE&RMSE&MAPE&MAE&RMSE&MAPE\cr 
    \hline
    2&1.31&2.72&2.72\%&1.62&\textbf{3.62}&3.60\%&1.90&4.34&4.42\% \cr
    4&\textbf{1.30}&\textbf{2.71}&\textbf{2.69}\%&\textbf{1.61}&\textbf{3.62}&\textbf{3.55}\%&\textbf{1.89}&\textbf{4.32}&\textbf{4.36}\% \cr
    6&\textbf{1.30}&2.74&2.72\%&1.63&3.71&3.67\%&1.92&4.45&4.59\% \cr
    \bottomrule
    \end{tabular}
    \label{hyper_bayB}
\end{table*}

\begin{table*}[!htb]
    \centering
    \begin{minipage}{.5\textwidth}
        \centering
         \caption{Impact of $M$ and $B$, PEMS03.}
    \vspace{-10pt}
    \begin{tabular}{c|ccc|c|ccc}
        \toprule  
        M&MAE&RMSE&MAPE&B&MAE&RMSE&MAPE \cr
    \hline
    3&14.95&25.36&15.18\%&2&14.92&25.11&15.03\% \cr
    5&\textbf{14.71}&\textbf{24.54}&\textbf{14.39}\%&4&\textbf{14.71}&\textbf{24.54}&\textbf{14.39}\% \cr
    7&14.82&25.23&14.51\%&6&14.80&24.73&14.45\% \cr
    \bottomrule
    \end{tabular}
    \label{hyper_03}
        
    \end{minipage}%
    \begin{minipage}{0.5\textwidth}
        \centering
     \caption{Impact of $M$ and $B$, PEMS04.}
    \vspace{-10pt}
    \begin{tabular}{c|ccc|c|ccc}
        \toprule  
        M&MAE&RMSE&MAPE&B&MAE&RMSE&MAPE \cr
    \hline
    3&19.40&30.64&13.76\%&2&19.39&30.74&13.67\% \cr
    5&\textbf{19.13}&\textbf{30.44}&\textbf{12.89}\%&4&\textbf{19.13}&\textbf{30.44}&\textbf{12.89}\% \cr
    7&19.32&30.69&13.18\%&6&19.11&30.53&12.90\% \cr
    \bottomrule
    \end{tabular}
    \label{hyper_04}
    \end{minipage}
    \vspace{-5pt}
\end{table*}

\begin{table*}[!htb]
    \centering
    \begin{minipage}{.5\textwidth}
        \centering
         \caption{Impact of $M$ and $B$, PEMS07.}
    \vspace{-10pt}
    \begin{tabular}{c|ccc|c|ccc}
        \toprule  
        M&MAE&RMSE&MAPE&B&MAE&RMSE&MAPE \cr
    \hline
    3&21.73&34.66&9.25\%&2&21.11&33.74&9.53\% \cr
    5&\textbf{20.93}&\textbf{33.69}&\textbf{8.90}\%&4&{20.93}&{33.69}&\textbf{8.90}\% \cr
    7&21.32&34.14&9.17\%&6&\textbf{20.62}&\textbf{33.30}&9.33\% \cr
    \bottomrule
    \end{tabular}
    \label{hyper_07}
        
    \end{minipage}%
    \begin{minipage}{0.5\textwidth}
        \centering
     \caption{Impact of $M$ and $B$, PEMS08.}
    \vspace{-10pt}
    \begin{tabular}{c|ccc|c|ccc}
        \toprule  
        M&MAE&RMSE&MAPE&B&MAE&RMSE&MAPE \cr
    \hline
    3&15.27&24.06&10.14\%&2&15.37&24.14&9.74\% \cr
    5&\textbf{14.82}&\textbf{23.64}&\textbf{9.51}\%&4&\textbf{14.82}&\textbf{23.64}&\textbf{9.51}\% \cr
    7&14.97&24.11&9.62\%&6&15.12&23.89&9.77\% \cr
    \bottomrule
    \end{tabular}
    \label{hyper_08}
    \end{minipage}
    \vspace{-5pt}
\end{table*}

\begin{table*}[!htbp]
\small
    \centering
    \caption{Impact of $M$ and $B$, Solar-Energy.}
    \vspace{-10pt}
    \begin{tabular}{c|cccc|c|cccc}
        \toprule  
        \multirow{2}{*}{M}&
        \multicolumn{2}{c}{3}&
        \multicolumn{2}{c}{24}&
        \multirow{2}{*}{B}&
        \multicolumn{2}{c}{3}&
        \multicolumn{2}{c}{24} \cr
        &RRSE&CORR&RRSE&CORR&&RRSE&CORR&RRSE&CORR \cr
    \hline
    3&{0.1786}&{0.9848}&{0.4150}&{0.9062}&2&0.1755&0.9846&0.4164&0.9040 \cr
    5&\textbf{0.1750}&\textbf{0.9855}&{0.4143}&\textbf{0.9085}&4&0.1750&0.9855&0.4143&0.9085 \cr
    7&0.1761&0.9849&\textbf{0.4136}&0.9073&6&\textbf{0.1692}&\textbf{0.9869}&\textbf{0.4117}&\textbf{0.9089} \cr
    \bottomrule
    \end{tabular}
    \label{hyper_solar}
    \vspace{-5pt}
\end{table*}

\begin{table*}[!htbp]
\small
    \centering
    \caption{Impact of $M$ and $B$, Electricity.}
    \vspace{-10pt}
    \begin{tabular}{c|cccc|c|cccc}
        \toprule  
        \multirow{2}{*}{M}&
        \multicolumn{2}{c}{3}&
        \multicolumn{2}{c}{24}&
        \multirow{2}{*}{B}&
        \multicolumn{2}{c}{3}&
        \multicolumn{2}{c}{24} \cr
        &RRSE&CORR&RRSE&CORR&&RRSE&CORR&RRSE&CORR \cr
    \hline
    3&0.0752&0.9441&\textbf{0.0942}&0.9223&2&0.0749&0.9438&0.0971&0.9235 \cr
    5&\textbf{0.0743}&\textbf{0.9477}&{0.0947}&\textbf{0.9239}&4&\textbf{0.0743}&\textbf{0.9477}&\textbf{0.0947}&\textbf{0.9239} \cr
    7&0.0755&0.9463&0.0949&0.9231&6&0.0748&0.9432&0.0958&0.9227 \cr
    \bottomrule
    \end{tabular}
    \label{hyper_elec}
\end{table*}

\begin{table*}[!htb]
    \centering
    \begin{minipage}{.5\textwidth}
        \centering
         \caption{Runtime and Parameters, METR-LA.}
    \begin{tabular}{cccc}
    \toprule  
    \multirow{2}{*}{Models}&
    {Training}&Inference&
    \multirow{2}{*}{Parameters} \cr
    &(s/epoch)&(ms/window)\cr 
    \midrule  
    DCRNN&304.2&8.58&372,353 \cr
    STGCN&16.8&7.32&119,176 \cr
    Graph WaveNet&46.7&0.17&309,400 \cr
    AGCRN&47.6&0.59&751,650 \cr
    MTGNN&39.6&0.43&405,452 \cr
    AutoSTG&453.3&1.60&509,048 \cr
    \emph{AutoCTS}&138.9&0.34&358,520
    \cr
    \bottomrule
    \end{tabular}
    \label{table_la}
        
    \end{minipage}%
    \begin{minipage}{0.5\textwidth}
        \centering
     \caption{Runtime and Parameters, PEMS-BAY.}
    \begin{tabular}{cccc}
    \toprule  
    \multirow{2}{*}{Models}&
    {Training}&Inference&
    \multirow{2}{*}{Parameters} \cr
    &(s/epoch)&(ms/window)\cr 
    \midrule  
    DCRNN&779.0&11.88&372,353 \cr
    STGCN&41.8&7.24&119,648 \cr
    Graph WaveNet&117.9&0.17&311,760 \cr
    AGCRN&95.3&0.80&752,830 \cr
    MTGNN&87.5&0.68&573,484 \cr
    AutoSTG&613.2&1.97&553,932 \cr
    \emph{AutoCTS}&310.4&0.35&395,984
    \cr
    \bottomrule
    \end{tabular}
    \label{table_bay}
    \end{minipage}
\end{table*}

\begin{table*}[!htb]
    \centering
    \begin{minipage}{.5\textwidth}
        \centering
         \caption{Runtime and Parameters, PEMS03.}
    \begin{tabular}{cccc}
    \toprule  
    \multirow{2}{*}{Models}&
    {Training}&Inference&
    \multirow{2}{*}{Parameters} \cr
    &(s/epoch)&(ms/window)\cr 
    \midrule  
    DCRNN&335.7&10.90&371,393 \cr
    STGCN&18.8&7.18&119,204 \cr
    Graph WaveNet&52.8&0.16&312,388 \cr
    AGCRN&45.7&0.94&749,320 \cr
    MTGNN&45.1&0.74&619,228 \cr
    \emph{AutoCTS}& 149.3&0.34&351,108 \cr
    \bottomrule
    \end{tabular}
    \label{table_03}
        
    \end{minipage}%
    \begin{minipage}{0.5\textwidth}
        \centering
     \caption{Runtime and Parameters, PEMS04.}
    \begin{tabular}{cccc}
    \toprule  
    \multirow{2}{*}{Models}&
    {Training}&Inference&
    \multirow{2}{*}{Parameters} \cr
    &(s/epoch)&(ms/window)\cr 
    \midrule  
    DCRNN&226.1&11.69&371,393 \cr
    STGCN&10.5&7.45&119,000 \cr
    Graph WaveNet&28.5&0.17&311,368 \cr
    AGCRN&21.9&0.74&748,810 \cr
    MTGNN&23.6&0.64&546,604 \cr
    \emph{AutoCTS}&80.3&0.34&367,048 \cr
    \bottomrule
    \end{tabular}
    \label{table_04}
    \end{minipage}
\end{table*}

\begin{table*}[!htb]
    \centering
    \begin{minipage}{.5\textwidth}
        \centering
         \caption{Runtime and Parameters, PEMS07.}
    \begin{tabular}{cccc}
    \toprule  
    \multirow{2}{*}{Models}&
    {Training}&Inference&
    \multirow{2}{*}{Parameters} \cr
    &(s/epoch)&(ms/window)\cr 
    \midrule  
    DCRNN&870.9&25.45&371,393 \cr
    STGCN&52.8&13.18&121,304 \cr
    Graph WaveNet&155.8&0.17&322,888 \cr
    AGCRN&144.8&2.76&754,570 \cr
    MTGNN&107.6&1.95&1,366,828 \cr
    \emph{AutoCTS}&381.2&0.35&407,112 \cr
    \bottomrule
    \end{tabular}
    \label{table_07}
        
    \end{minipage}%
    \begin{minipage}{0.5\textwidth}
        \centering
     \caption{Runtime and Parameters, PEMS08.}
    \begin{tabular}{cccc}
    \toprule  
    \multirow{2}{*}{Models}&
    {Training}&Inference&
    \multirow{2}{*}{Parameters} \cr
    &(s/epoch)&(ms/window)\cr 
    \midrule  
    DCRNN&157.7&7.61&371,393 \cr
    STGCN&6.4&4.50&118,452 \cr
    Graph WaveNet&18.7&0.16&308,628 \cr
    AGCRN&16.6&0.53&150,112 \cr
    MTGNN&14.1&0.35&351,516 \cr
    \emph{AutoCTS}&50.3&0.34&357,236 \cr
    \bottomrule
    \end{tabular}
    \label{table_08}
    \end{minipage}
\end{table*}

\begin{table*}[!htb]
    \centering
    \begin{minipage}{.5\textwidth}
        \centering
         \caption{Runtime and Parameters, Solar-Energy.}
    \begin{tabular}{cccc}
    \toprule  
    \multirow{2}{*}{Models}&
    {Training}&Inference&
    \multirow{2}{*}{Parameters} \cr
    &(s/epoch)&(ms/window)\cr  
    \midrule  
    LSTNet&29.2&0.06&174,807 \cr
    TPA-LSTM&206.8&11.78&122,295 \cr
    MTGNN&218.3&1.82&347,665 \cr
    \emph{AutoCTS}&746.1&1.06&318,277 \cr
    \bottomrule
    \end{tabular}
    \label{table_solar}
        
    \end{minipage}%
    \begin{minipage}{0.5\textwidth}
        \centering
     \caption{Runtime and Parameters, Electricity.}
    \begin{tabular}{cccc}
    \toprule  
    \multirow{2}{*}{Models}&
    {Training}&Inference&
    \multirow{2}{*}{Parameters} \cr
    &(s/epoch)&(ms/window)\cr 
    \midrule  
    LSTNet&12.1&0.05&325,871 \cr
    TPA-LSTM&252.2&30.94&193,135 \cr
    MTGNN&258.9&4.51&362,385 \cr
    \emph{AutoCTS}&856.0&3.50&321,957 \cr
    \bottomrule
    \end{tabular}
    \label{table_elec}
    \end{minipage}
    \vspace{90pt}
\end{table*}

\begin{table*}[h]
\small
    \centering
    \caption{Transferability: Transferred Model is {Searched} on PEMS03, \emph{AutoCTS} is {Searched} on METR-LA or PEMS-BAY.}
    \begin{threeparttable} 
    \begin{tabular}{ccccc|ccc|ccc}
        \hline
        \toprule  
        \multirow{2}{*}{Data}&
        \multirow{2}{*}{Models}&
        
        \multicolumn{3}{c}{15 min}&
        \multicolumn{3}{c}{30 min}&
        \multicolumn{3}{c}{60 min} \cr  
    & &MAE&RMSE&MAPE&MAE&RMSE&MAPE&MAE&RMSE&MAPE\cr  
    \midrule  
    \multirow{2}{*}{{METR-LA}}& Transferred Model& 2.72&\textbf{5.11}&6.90\%&3.08&\textbf{6.09}&8.28\%&3.50&\textbf{7.12}&10.02\% \cr
    & \emph{AutoCTS}& \textbf{2.67}&\textbf{5.11}&\textbf{6.80}\%&\textbf{3.05}&{6.11}&\textbf{8.15}\%&\textbf{3.47}&{7.14}&\textbf{9.81}\% \cr
    \hline
    \multirow{2}{*}{{PEMS-BAY}}& Transferred Model& \textbf{1.30}&2.73&2.73\%&1.62&3.63&3.63\%&1.90&4.35&4.51\% \cr
    & \emph{AutoCTS}& \textbf{1.30}&\textbf{2.71}&\textbf{2.69}\%&\textbf{1.61}&\textbf{3.62}&\textbf{3.55}\%&\textbf{1.89}&\textbf{4.32}&\textbf{4.36}\% \cr
    \bottomrule
    \end{tabular}
    \end{threeparttable}
    \label{table9}
\end{table*}

\begin{table*}[htbp]
\small
    \centering
    \caption{Impact of the number of incoming edges per node, METR-LA.}
    \begin{tabular}{c|ccccccccc|c}
        \toprule  
        \multirow{2}{*}{\#  Edges}&
        \multicolumn{3}{c}{15 min}&
        \multicolumn{3}{c}{30 min}&
        \multicolumn{3}{c}{60 min}&
        \multirow{2}{*}{Training (s/epoch)} \cr
    &MAE&RMSE&MAPE&MAE&RMSE&MAPE&MAE&RMSE&MAPE\cr 
    \hline
    2&\textbf{2.67}&{5.11}&\textbf{6.80}\%&\textbf{3.05}&{6.11}&\textbf{8.15}\%&\textbf{3.47}&\textbf{7.14}&{9.81}\%&\textbf{138.9} \cr
    3&2.68&\textbf{5.09}&6.86\%&\textbf{3.05}&\textbf{6.07}&8.16\%&3.50&7.18&\textbf{9.75}\%&188.6 \cr
    \bottomrule
    \end{tabular}
    \label{edges_la}
\end{table*}

\begin{table}[htbp]
\small
    \centering
    \caption{Impact of the number of edges per node, PEMS03.}
    \begin{tabular}{c|cccc}
        \toprule  
        \# Edges&MAE&RMSE&MAPE&Training (s/epoch) \cr
    \hline
    2&{14.71}&{24.54}&\textbf{14.39}\%&\textbf{149.3} \cr
    3&\textbf{14.58}&\textbf{24.20}&{15.40}\%&204.0 \cr
    \bottomrule
    \end{tabular}
    \label{edges_03}
\end{table}

\begin{table}[!htbp]
\small
    \centering
    \caption{Categorization of Human Designed ST-blocks.}
    \begin{tabular}{l|c|c|c}
        \hline
        &CNN&RNN&Attention \cr
        \hline
    GCN&\cite{yu2018spatio,guo2019attention,Wu2019graph,huang2020lsgcn,diao2019dynamic,fang2019gstnet,wu2020connecting}&\cite{li2018dcrnn_traffic,bai2020adaptive,chen2020multi,DBLP:conf/icde/Hu0GJX20}&\cite{guo2019attention} \cr
    Attention&\cite{guo2019attention}&None&\cite{zheng2020gman,xu2020spatial} \cr
    \hline
    \end{tabular}
    \label{related-work}
\end{table}

In particular, we compare \emph{AutoCTS} with the following variants: (1) w/o design principles: this variant does not follow the proposed two principles for selecting a compact set of S/T operators. Rather, it includes all operators in Table~\ref{table1}. 
(2) w/o temperature: it does not use the temperature parameter $\tau$ to reduce the gap between the micro-DAG and the derived ST-block. 
(3) w/o macro search: this variant only searches for a single optimal ST-block and then sequentially stacks ST-blocks with residual connections to build an ST-backbone. 
(4) macro only: 
it employs four existing human designed ST-blocks as the atomic search  units and only searches for novel ST-backbones. The selected ST-blocks come from STGCN~\cite{yu2018spatio}, DCRNN~\cite{li2018dcrnn_traffic}, Graph WaveNet~\cite{Wu2019graph}, and MTGNN~\cite{wu2020connecting}. 
We consider (1) the accuracy of the models that are identified by the different variants, and (2) the runtime in GPU hours that it takes to identify the models. 

Tables~\ref{table_la_ablation} to ~\ref{table_elec_ablation} show 
that: 
(1) \emph{AutoCTS} achieves better accuracy than its variant w/o design principles, and costs much less GPU hours for architecture search. This demonstrates the effectiveness of the proposed principles for selecting a compact and complementary S/T operators from Table~\ref{table1}. 
(2) The proposed temperature parameter helps reduce the gap and find more accurate models with similar GPU hours, except in the case of only one dataset, Solar-Energy.
(3) Disabling the macro search and search for the stacking of homogeneous ST-backbone lowers the performance without significantly decreasing the searching time -- the only exception is the METR-LA dataset, where the accuracies when using and no using the macro search are similar. This suggests our joint search space and strategy is effective and efficient. 
(4) \emph{AutoCTS} significantly outperforms the macro only variant, which justifying that S/T operators are more suitable to be used as the atomic search units in the search space than manually-designed ST-blocks. %
Although the macro only variant is very efficient, due to its small search space, it is unappealing as many human designed models, such as Graph WaveNet, AGCRN, and MTGNN, outperform it.  

\subsubsection{{Parameter Sensitivity Analysis}}
We proceed to evaluate the impact of key hyperparameters in \emph{AutoCTS}, including $M$, i.e., the number of nodes in an ST-block in the micro search space, and $B$, i.e., the number of ST-blocks $B$ in the macro search space{, and $Edge$, i.e., the number of incoming edges per node in the derived ST-block.
We use $B=4$, $M=5$, and $Edge=2$ as default values. We then  
vary $M$ among \{3, 5, 7\}, 
vary $S$ among \{2, 4, 6\}, 
and vary $Edge$ among \{2, 3\}, while keeping the rest to their default values. 

}


%
Tables~\ref{hyper_laM} to ~\ref{hyper_elec} show that \emph{AutoCTS} achieves the best accuracy under $M=5$ and $B=4$, except on PEMS07 and Solar-Energy, where the results are not optimal but are still comparable to the best setting. 
Decreasing $M$ or $B$ reduces the expressiveness of \emph{AutoCTS} and thus the accuracy of the automatically identified models. 
%
A larger $M$ or $B$ increases the complexity of the micro and macro search space, resulting in potentially more overfitting problems when the training data is not abundant. Thus, it slightly degrades the accuracy. {In addition, larger $M$ or $B$ may lead to models that use significantly more parameters than the baseline models. }

{
Tables \ref{edges_la} and \ref{edges_03} show minimal accuracy improvements when $Edge$ increases from 2 to 3, while the training time (s/epoch) 
shows a clear increase. This suggests that using 2 edges per node yields sufficiently complex internal topologies for ST-blocks and avoids introducing too many parameters and thus maintaining good efficiency.
}

\subsubsection{Case Study}

\begin{figure*}[htbp]
\center
\subfigure[ST-block 1]{
\begin{minipage}[c]{0.17\linewidth} 
\centering
\includegraphics[width=\linewidth]{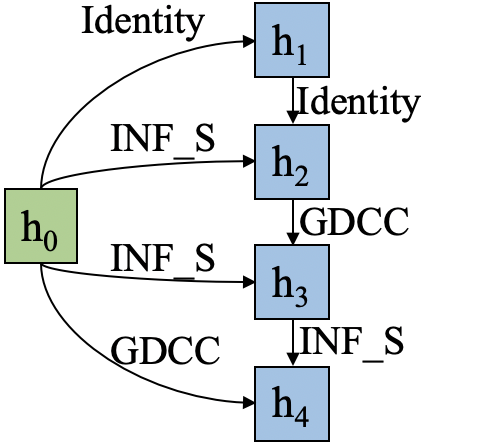} 
\label{b1}
\end{minipage}
}
\subfigure[ST-block 2]{
\begin{minipage}[c]{0.17\linewidth}
\centering
\includegraphics[width=\linewidth]{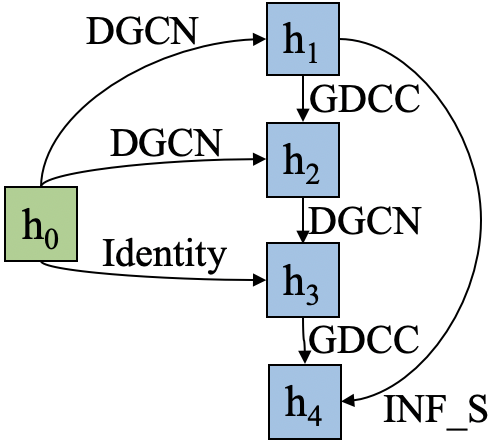}
\label{b2}
\end{minipage}
}
\subfigure[ST-block 3]{
\begin{minipage}[c]{0.17\linewidth}
\centering
\includegraphics[width=\linewidth]{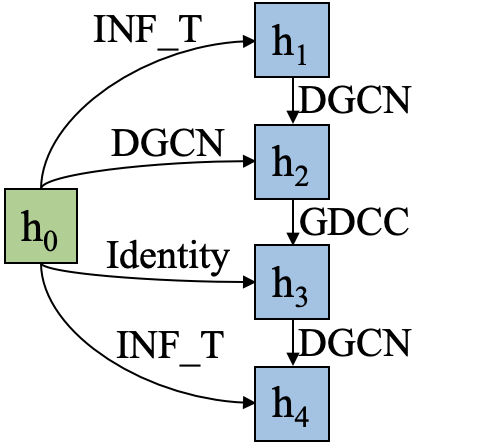}
\label{b3}
\end{minipage}
}
\subfigure[ST-block 4]{
\begin{minipage}[c]{0.17\linewidth}
\centering
\includegraphics[width=\linewidth]{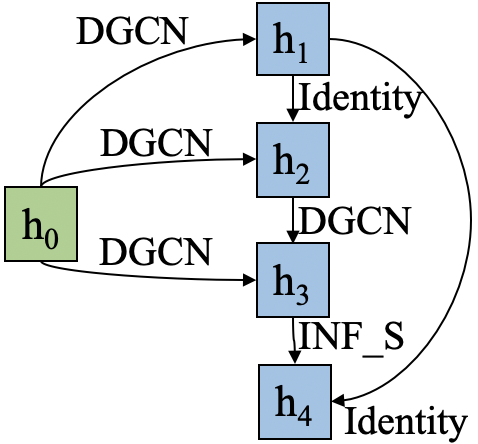}
\label{b4}
\end{minipage}
}
\subfigure[Forecasting Model]{
\begin{minipage}[c]{0.21\linewidth}
\centering
\includegraphics[width=\linewidth]{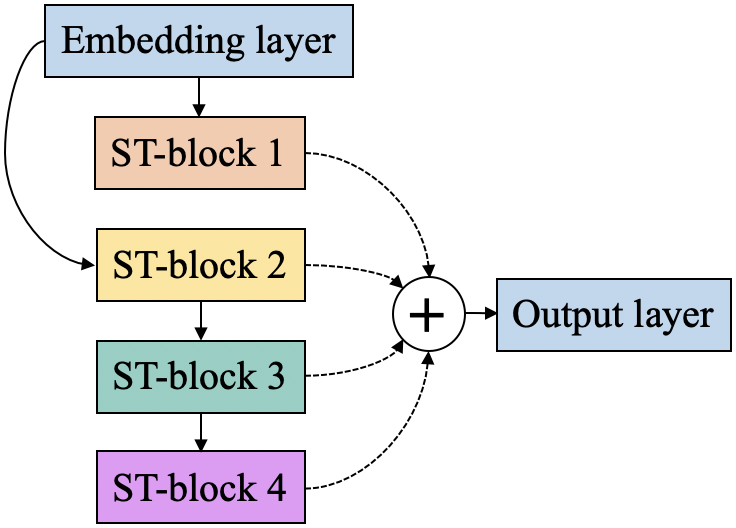}
\label{searched-macro}
\end{minipage}
}

\caption{The Automatically Searched Forecasting Model on the PEMS03 Dataset. 
}

\label{casestudy}
\end{figure*}

We show the architecture of the forecasting model on  PEMS03 in Figure~\ref{casestudy}. 
As Figures~\ref{b1}, ~\ref{b2}, ~\ref{b3} and ~\ref{b4} show, each ST-block has a distinct internal architecture. In particular, the four ST-blocks contain all S/T operators in the micro search space, including 5 GDCC, 2 INF-T, 5 INF-S and 10 DGCN. This indicates the effectiveness of the proposed micro search space. 
The ST-backbone consists of the four heterogeneous ST-blocks, which are assembled by diverse topologies. This justifies the needs of enabling topologically flexible, heterogeneous ST-backbone, which most existing models fail to support.  

%
%

%


\subsubsection{Transferability}

{Since manually designed forecasting models are often applied to different datasets, it is pertinent to investigate the transferability of forecasting models learned by \emph{AutoCTS} to assess how such models compare with traditional, manually designed models.
To this end, we consider a ``Transferred Model'' that is identified automatically by \emph{AutoCTS} on the PEMS03 dataset, as shown in Figure ~\ref{casestudy}. We apply this model to make forecasts on datasets METR-LA and PEMS-BAY. 
%
The results are shown in Table~\ref{table9}, where  \emph{AutoCTS} denotes the automatically identified model on METR-LA or PEMS-BAY. 
%
%
%
The transferred model achieves competitive accuracy on METR-LA and
PEMS-BAY. Although the transferred model is not as good as the
model that is directly learned by \emph{AutoCTS} on the specific dataset, it
is able to outperform the baselines on most metrics, especially in the case of
PEMS-BAY (cf. Table~\ref{table4}). This is evidence that \emph{AutoCTS} is able to
produce effective and transferable forecasting models.
}

\subsubsection{Runtime, Memory Costs \& Number of Parameters}
For the architecture search phase, we consider 
the runtime and memory that \emph{AutoCTS} takes. 
Table~\ref{memory} shows that the search time varies from 12.34 to 163.21 GPU hours across datasets, depending on the number of time series/nodes, the total number of timestamps, 
and the length of the input time window. The searching process takes up to ca. 36 GB memory, which can fit into the memory of a single modern GPU. 


To understand how models identified by \emph{AutoCTS} and baseline models compare in terms of time and space, we report the training time (seconds per epoch), inference time (milliseconds per window), and total number of parameters in Tables~\ref{table_la} to ~\ref{table_elec}. 
First, in terms of training time,
\emph{AutoCTS} is faster than DCRNN and AutoSTG but slower than other models. It's because \emph{AutoCTS} includes operators from both the CNN and Attention families, where Attention operators are less efficient than CNN operators. Besides, \emph{AutoCTS} is much deeper than DCRNN and AGCRN. DCRNN and AutoSTG are slow because the former sequentially produces predictions at all time steps, and the latter employs meta-learning to learn the network weights.


Second, in terms of inference time, all methods are able to make near-instantaneous predictions when new data arrives and are thus able to support online prediction in streaming settings. 
Third, in terms of the number of parameters, AutoCTS often uses fewer parameters than do MTGNN, AGCRN, and AutoSTG, uses more parameters than does STGCN (whose accuracy is among the worst), and is comparable to the other methods.


\section{Related work}
\label{Sec 2}
We categorize existing studies on CTS forecasting into two categories---manually designed models vs. automated designed models. 

\noindent \textbf{Manually Designed Models. }
%
Recent deep learning models achieve the state-of-the-art accuracy in correlated time series forecasting. 
%
Such models rely on different types of human designed ST-blocks that capture both temporal dependencies and spatial correlations. 
Table~\ref{related-work} summarizes the ST-blocks in the literature according to two dimensions---temporal dependencies modeling (including the CNN, RNN and attention families) vs. spatial correlation modeling (including the GCN and attention families). 


\noindent \textbf{Automatically Designed Models. } 
Neural Architecture Search (NAS) has been employed to automatically design neural architectures for the many tasks. Existing NAS methods can be divided into evolutionary algorithm based~\cite{2019The}, reinforcement learning based~\cite{pham2018efficient}, performance predictor based~\cite{li2020neural} and gradient-based methods~\cite{liu2018darts}. \emph{AutoCTS} is a gradient-based method due to its high efficiency.
Despite of great success in 
computer vision~\cite{pham2018efficient,chen2018searching}, natural language processing~\cite{2019The}, and AutoML systems~\cite{zogaj2021doing,li2021volcanoml}, little effort has been devoted to time series forecasting. 

%
AutoST~\cite{li2020autost} is proposed for spatio-temporal prediction, where the time series are from a uniform grid. The values at each timestamp are considered as an image, and 
then a search space that only contains convolution operators is proposed. AutoST does not apply in our setting, where time series are not necessarily from a uniform grid, making the image modeling inapplicable. 
%
AutoSTG~\cite{pan2019urban} considers correlated time series forecasting. However, it differs from \emph{AutoCTS} in the following perspective. (1) AutoSTG only considers one T-operator and one S-operator, i.e., 1D convolution and Diffusion GCN. In contrast, we propose two principles to select the most effective and efficient operators from diverse families. (2) AutoSTG only designs ST-blocks, whereas we search jointly both ST-blocks and the ST-backbone. (3) AutoSTG relies on additional information of the graph to enable meta-learning to learn network weight $w$. In contrast, \emph{AutoCTS} does not rely on such additional information but purely on the time series themselves. Thus, \emph{AutoCTS} has a wider application scope.  

\section{Conclusion}
\label{Sec 6}
We present \emph{AutoCTS}, a 
framework that is able to automatically learn a neural network  model for correlated time series forecasting. In particular, we design a micro search space with a compact set of S/T operators 
to find 
novel ST-blocks. In addition, we design a macro search space to identify the topology among heterogeneous ST-blocks to construct novel ST-backbones. 
%
Extensive experiments on eight commonly used correlated time series forecasting datasets justify the design choices of \emph{AutoCTS}. 
As future work, it is of interest to include model efficiency as an additional criterion into the search strategy to automatically identify both accurate and efficient models and to employ parallel computing techniques~\cite{DBLP:conf/waim/YuanSWYZY10} to make the search more efficient. It is also of interest to extend \emph{AutoCTS} to other analytics tasks, such as outlier detection~\cite{tungicde2022} and trajectory analytics~\cite{DBLP:conf/ijcai/YangGHT021}. 


\begin{acks}
This work was partially supported by Independent Research Fund Denmark under agreements 8022-00246B and 8048-00038B, the VILLUM FONDEN under agreements 34328 and 40567, and the Innovation Fund Denmark centre, DIREC.
\end{acks}


\bibliographystyle{ACM-Reference-Format}
\bibliography{sample}

\end{document}